%% file: main.tex
\documentclass{article}

\PassOptionsToPackage{numbers, compress}{natbib}
\usepackage[main, final]{neurips_2026}
\usepackage[utf8]{inputenc} 
\usepackage[T1]{fontenc}    
\usepackage{hyperref}       
\usepackage{url}            
\usepackage{booktabs}       
\usepackage{amsfonts}       
\usepackage{nicefrac}       
\usepackage{microtype}      
\usepackage{xcolor}         
\usepackage{amsmath}
\usepackage{amssymb}
\usepackage{amsthm}
\usepackage{graphicx}
\usepackage{cleveref}
\usepackage{placeins}
\usepackage{pifont}
\usepackage{multirow}
\usepackage{subcaption}
\usepackage{caption}

\usepackage[acronym]{glossaries}
\newacronym{ei}{EI}{Expected Improvement}
\newacronym{rkhs}{RKHS}{Reproducing Kernel Hilbert Space}
\newacronym{rbf}{RBF}{Radial Basis Function}
\newacronym{bo}{BO}{Bayesian Optimization}
\newacronym{fbo}{FBO}{Functional Bayesian Optimization}
\newacronym{gp}{GP}{Gaussian Process}
\newacronym{lhs}{LHS}{Latin Hypercube Sampling}
\newacronym{mse}{MSE}{Mean Squared Error}
\newacronym{mlp}{MLP}{Multi Layer Perceptron}
\newacronym{rl}{RL}{Reinforcement Learning}
\newacronym{ci}{CI}{Confidence Interval}
\newacronym{kmp}{KMP}{Kernel Matching Pursuit}

\glsdisablehyper

\newcommand{\wycoff}{\textsc{L0MO}}
\newcommand{\vien}{\textsc{Vien}~\cite{vien2018bayesian}}
\newcommand{\vellanky}{\textsc{Vellanki}~\cite{vellanki2019bayesian}}
\newcommand{\kundu}{\textsc{Kundu}~\cite{kundu2025active}}
\newcommand{\shilton}{\textsc{Shilton}~\cite{shilton2020sequential}}
\newcommand{\random}{\textsc{Random}}

\newcommand{\N}{\mathbb{N}}
\newcommand{\R}{\mathbb{R}}
\newcommand{\U}{\mathcal{U}} 
\newcommand{\D}{\mathcal{D}} 
\newcommand{\F}{\mathcal{F}}

\newcommand{\Hs}{\mathcal{H}}
\newcommand{\transposed}{^\intercal}
\newcommand{\norm}[1]{\lVert #1 \rVert}
\newcommand{\inner}[2]{\langle #1, #2 \rangle}
\newcommand{\partialover}[1]{\frac{\partial}{\partial #1}}
\newcommand{\acq}{\mathcal{I}}

\newcommand{\kernelH}{k}
\newcommand{\kernelGP}{\kappa}
\newcommand{\vu}{u}

\title{
Bayesian Optimization on Function Spaces via \\ Sparse RKHS Manifolds
}

\author{%
  \textbf{Davide Sartor} \\
  Department of Mathematics and Statistics \\
  University of Massachusetts Amherst \\
  \texttt{dsartor@umass.edu} \\
  \And
  \textbf{Meghan E.\ Huber} \\
  Department of Mechanical \\
  and Industrial Engineering \\
  University of Massachusetts Amherst \\
  \texttt{mehuber@umass.edu} \\
  \AND
  \textbf{Donghyun Kim} \\
  Manning College of Information \\
  and Computer Sciences \\
  University of Massachusetts Amherst \\
  \texttt{donghyunkim@cs.umass.edu} \\
  \And
  \textbf{Nathan Wycoff} \\
  Department of Mathematics and Statistics \\
  University of Massachusetts Amherst \\
  \texttt{nwycoff@umass.edu} \\
}

\begin{document}

\maketitle

\begin{abstract}
\gls{bo} has become an established methodology for minimizing black-box functions of a vector input.
Often, however, this parameter vector arises from the discretization of an inherently functional relationship. 
Several recent articles have considered the \gls{fbo} setting, in which the variable to be optimized is not a member of a finite dimensional vector space, but rather an infinite dimensional function space.
In this work, we propose $L^0$ Manifold Optimization (\wycoff), a simple approach to \gls{fbo} which searches the subset of a \gls{rkhs} consisting of functions with a sparse representation in the kernel functions, optimizing both the kernel locations and their coefficients.
We discuss in detail the relationship between our method and existing ones, providing a unifying lens through which to view prior works.
To assess our method against the state of the art, we conduct an extensive computational study, and along the way develop a novel set of benchmark test functions which port standard finite-dimensional ones to the infinite dimensional domain.
Our experiments demonstrate that, on balance, the proposed method achieves superior performance across a wide range of test benchmarks.
\end{abstract}

\section{Introduction}

\begin{figure}[t]
  \centering
  \includegraphics[width=\textwidth]{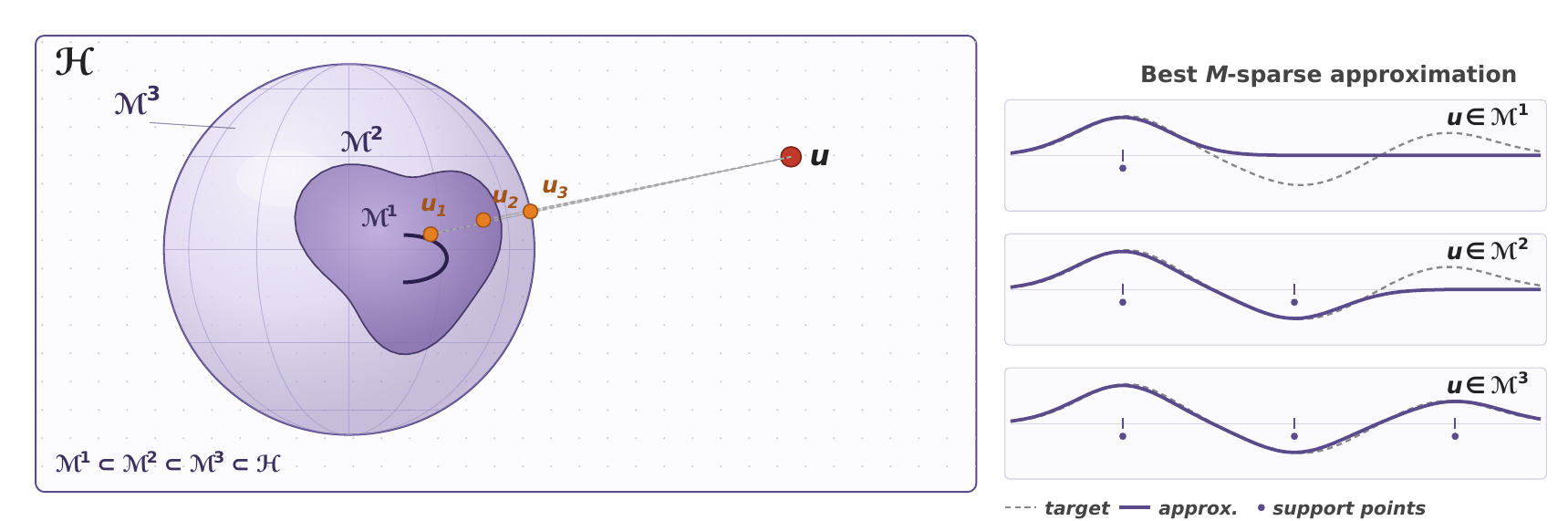}
  \caption{A schematic view inside $\mathcal{H}$, illustrating our approximation strategy.
  The sparse manifolds form a stratified set of increasing dimension, depicted as a curve, a surface patch, and a ball.
  For a target $u \in \mathcal{H}$ that does not admit a finite sparse representation, the best $m$-sparse approximation $u_m \in \bigcup_{i \le m} \mathcal{M}^i$ moves closer to $u$ as $m$ grows.
  }
  \label{fig:natgrad}
  \vspace{-0.1in}
\end{figure}

Many critical applied tasks across chemistry \citep{shields2021bayesian,wang2022bayesian}, materials science \citep{frazier2015materials,pedersen2021bayesian,todorovic2019bayesian} and biomechanics \cite{kim2026validation}, among other areas, can be formulated as \emph{black-box optimization} problems of the form:
\begin{equation}
    \min_{\vu \in \U}
    f(\vu) \,,
\end{equation}
where the domain $\U$ of $f$ is usually a compact subset of $\R^P$, most commonly the hypercube $[0,1]^P$.
A leading strategy for addressing these problems is \gls{bo} \citep{jones1998efficient,frazier2018tutorial,garnett2023bayesian}, which involves reasoning probabilistically about the location of the global minimum given a set of observed input-output relationships $(\vu_1,f(\vu_1)),\ldots,(\vu_N,f(\vu_N))$.
Given a small initial sample and a total budget $B$, the goal is to choose points one-by-one so as to find the point $\vu$ with the lowest possible $f(\vu)$.

The finite dimensional setting is appropriate when $f$ depends on a finite set of clearly defined parameters.
Frequently, however, the parameters are actually a discretization of an intrinsically functional quantity.
For example, the classical \textit{wing weight function} \cite[Ch.~1]{forrester2008} is typically viewed as an 11 dimensional parametric function. 
But these parameters encode the shape of a wing, and the space of possible designs actually consists of the in-principle infinite dimensional space of wing shapes.

Many other scientists have effected similar discretizations: subject area expertise is used to tame the infinite dimensional nature of the problem and therefore make it amenable to the available \gls{bo} methodology and software environment.
However, in so doing, they have also restricted the solution space before the campaign has begun: not only is it possible that some slight improvement over the given discretization could be obtained if a flexible functional form were allowed, this conversion to a finite-dimensional setting means that we restrict \gls{bo} to learning solutions already conceivable by the subject matter expert.
But this rules out the most exciting possibility: that there is some function of a radically different nature than the state of the art which yields unimagined decrease in $f$.


This explains the nascent interest in the research community in \gls{fbo}.
There, the search space $\U$ is a subset of some space $\Hs$ of $\D\to\mathbb{R}$ functions.
For the sake of clarity we will use the term \emph{function} to refer to the input $\vu\in\mathcal{U}$, and the term \emph{functional} to refer to $f$.
We are aware of only a few studies which directly consider \gls{fbo}, and there remains much work to be done in studying this category of problems.
\cite{shilton2020sequential} and \cite{kundu2025active} search within random linear subspaces, yielding a finite dimensional problem in effect.
\cite{vellanki2019bayesian} search the space of $n$-th degree Bernstein polynomials, allowing the enforcement of shape priors. They propose a strategy for choosing the polynomial degree, but it is not clear how to scale the approach beyond one-dimensional functions.
\cite{vien2018bayesian} show that starting with some initial design of functions and then iteratively conducting Bayesian acquisition is tractable, insofar as the trajectory obtained by gradient descent in the function space stays within the span of the kernel functions defining the initial design. 
These methods all assume that a good function can be found within the initially specified linear subspace.

In this article, we will propose to solve the acquisition subproblem of the form: 
\begin{equation}
    \max_{w\in\mathcal{W}\subseteq \R^l}
    \acq \left(\phi(w) \right) \,,
\end{equation}
where $\acq$ is the acquisition function and $\phi: \R^l \to \Hs$ is a smooth parameterization of a stratified set in $\Hs$, obtained as the union of manifolds $\mathcal{M}$ of dimension at most $l$. 
For a given $l$, this allows a much richer set of candidate functions than existing fixed-basis method: our approach allows the support points themselves to vary.
This also allows for straightforward enforcement of additional desirable properties such as boundedness and sparsity, while maintaining compatibility with the computational tools used for finite-dimensional \gls{bo} such as multi-start Quasi-Newton search \cite[Chapter 7]{gramacy2020surrogates}.

Our contributions can be summarized as follows:
\begin{enumerate}
    \item We provide an extensive discussion of disparate existing methods for \gls{fbo}, providing a unifying lens and showing that \cite{vien2018bayesian} can be interpreted as natural gradient descent in a finite-dimensional parameter space.
    \item We introduce a ridge function \cite{pinkus2015ridge} approach to defining \gls{fbo} problems from finite dimensional \gls{bo} problems, which allows the use of test problems familiar to the \gls{bo} community.
    We release an open source python implementation of $48+$ popular \gls{bo} benchmarks and their extension to the functional domain,
    as well as several new realistic \gls{fbo} problems.
    \item We introduce a new method for \gls{fbo}, which we call \wycoff{}. We conduct an extensive simulation campaign across \gls{fbo} hyperparameters and a range of synthetic and real test problems, finding that the proposed method performs best overall.
    We also release the first open source codebase for \gls{fbo} at \url{https://github.com/davidesartor/bofus}.
\end{enumerate}

\section{Background}

\paragraph{Gaussian Processes and Bayesian Optimization}

Gaussian processes \citep{rnw} are a rule for assigning to every finite subset of points $\mathcal{X}$ in a domain $\D$ a joint normal distribution $\mathcal{N}(\mu(\mathcal{X}), \Sigma(\mathcal{X}))$.
The mean function $\mu$ is typically taken to be a simple polynomial or even a constant (we do the latter in our numerical experiments), and determining $\Sigma$ is the subject of most modeling decisions.
Each element $\Sigma_{i,j}$ is given by a \emph{kernel function} $\kernelGP(x,x')$; the set of functions yielding valid covariance matrices are the \emph{positive definite functions}.
$\kernelGP$ is often given as a monotonically decreasing function of some measure of dissimilarity between its two arguments.
In the most common application of GPs to vector inputs, the \gls{rbf} kernel $\kernelGP_{\textrm{RBF}}(x,x') = \sigma^2 e^{-\frac{1}{2\ell}\Vert x-x'\Vert_2^2}$ is used.
Note the presence of the hyperparameters $\sigma^2$ and $\ell$, which must be learnt, most commonly via numerical likelihood maximization, and this is also the approach taken in this article.
A key aspect of Gaussian processes that makes them practical is the tractability of their conditional distributions.
If we observe the value of a GP realization at some set of points $\mathcal{X}_{\textrm{Train}}$, then the posterior distribution is again a Gaussian process with updated mean and covariance functions, straightforwardly computable via the multivariate normal conditioning equations.

The capability to produce a full posterior over any new candidate point $x^*$ given an observation set $\mathcal{X}, f(\mathcal{X})$ undergirds the suitability of a \gls{gp} to perform sequential design.
In \gls{bo}, this proceeds by defining an \emph{acquisition function} $\acq$ which measures the desirability of $x^*$ in light of the goal of minimization and the information revealed by the data so far observed.
A popular acquisition function, and the one we use in this work is \gls{ei}:
\begin{equation}
    \acq_{\textrm{EI}}(\vu \mid \mathcal{GP}) = \sigma_{\mathcal{GP}}(\vu) \!\big( \gamma(\vu) \Phi(\gamma(\vu)) + \varphi(\gamma(\vu)) \big), \quad \gamma(\vu) = \frac{y^* - \mu_{\mathcal{GP}}(\vu)}{\sigma_{\mathcal{GP}}(\vu)},
\end{equation}
where $y^*$ is the current best observation and $\Phi$ and $\varphi$ are the standard normal CDF and PDF.
The next point to acquire is the one which maximizes the acquisition function.
Numerically resolving this is no small task, and typically involves a multi-start quasi-Newton search.

\paragraph{Gaussian Processes on Function Spaces}
Given a kernel function $\kernelH$, its \gls{rkhs} is the closure of all functions of the form
$\vu = \sum_{i=1}^n \alpha_i \kernelH(\cdot, x_i)$
for any $n$, equipped with an inner product which privileges these kernel functions.
In other words, given a \textit{support set} of functions $\F:=\{\kernelH(\cdot,x) : x \in \D\}$, the \gls{rkhs} $\Hs$ is $\overline{\textrm{span}(\F)}$.
For any two functions expressible as a finite linear combination of support functions,
$\vu = \sum_{i=1}^n \alpha_i \kernelH(\cdot, x_i)$
and
$\vu' = \sum_{j=1}^m \alpha'_j \kernelH(\cdot, x'_j)$,
we have that
$
{\inner{\vu}{\vu'}_\Hs = \sum_{i,j} \alpha_i \alpha_j' \kernelH(x_i,x'_j)},
$
and extend $\langle\cdot,\cdot\rangle_{\Hs}$ to the whole space by continuity.

As GPs rely on the domain on which they are fit only via $\kernelGP$, they are applicable to a wide range of mathematical objects, including functions.
In fact, most kernels can be decomposed into a dissimilarity measure, composed with a \emph{profile} function. In particular, given some Hilbert space of functions $\Hs$ with norm $\norm{\cdot}_\Hs$, we can populate the covariance matrix entry between two functions $\vu_i,\vu_j\in\Hs$ via $\Sigma_{i,j} = p(\norm{\vu_i - \vu_j}_\Hs)$ for a given profile function $p$.
In general, this requires approximating the norm on the function space \citep{muehlenstaedt2017computer,betancourt2020fungp,damiano2022automatic}.
Following \cite{vien2018bayesian}, we use an \gls{rkhs} as our function space.
Since we are in the context of a designed experiment rather than observational data, we can afford to choose functions which lead to an easily computed $\norm{\cdot}_\Hs$.
In particular, if two functions $\vu,\vu' \in \Hs$ are expressible with a finite linear combination of elements of $\F$ we have that 
\begin{equation}
\norm{\vu - \vu'}_\Hs^2 = \sum_{i,j} \alpha_i\alpha_j \kernelH(x_i,x_j) - 2\sum_{i,j} \alpha_i\alpha_j' \kernelH(x_i,x_j') + \sum_{i,j} \alpha_i'\alpha_j' \kernelH(x_i',x_j'),
\end{equation}
which can be computed efficiently if neither function consists of overly many basis points.

\gls{rkhs}'s and Gaussian processes are intimately linked \citep{kanagawa2018gaussian}: for example, the posterior mean of a Gaussian process regression lives in the \gls{rkhs} induced by the kernel used for the GP.
It will be important to avoid confusion between $\kernelGP: \Hs \times\Hs \to \R$, the kernel of the \gls{gp} used as surrogate model, which accepts two functions in $\Hs$ as input and outputs an element of the covariance of the Gaussian process, and $\kernelH:\R^d\times\R^d\to\R$, the kernel of the \gls{rkhs} that represents the functional domain $\D$ of interest, and which is used to construct our candidate functions $u$.

In this work, it will be useful to define an extension of the $\ell^0$ ``norm"
\footnote{As it clearly fails to obey positive homogeneity, this is not technically a norm. The same will be true of our extension.}
for vectors, which counts its argument's nonzero coefficients, to functions in an RKHS generated by $\F$:
\begin{equation}
    \Vert \vu \Vert_0 = \inf\Big\{ m \in \N : \exists\, \alpha_1,\ldots,\alpha_m \in \R,\ x_1,\ldots,x_m \in \D \text{ such that } \vu = \textstyle\sum_{i=1}^m \alpha_i \kernelH(\cdot, x_i) \Big\},
\end{equation}
with $\inf\emptyset = \infty$.
We refer to this as the $L^0$ norm.
The functions of finite $L^0$ norm form the pre-\gls{rkhs} $\Hs_0 = \textrm{span}(\F)$, which is dense in $\Hs$ by construction.
Notice that while the vector constraint set $\{w \in \R^P : \Vert w \Vert_0\leq m\}$ is a union of coordinate subspaces that meet only at low-sparsity points, in $\Hs$ the support locations $x_i$ can move continuously, so that any two functions with $\Vert \vu \Vert_0 \leq m$ are connected by a path within the set; this facilitates optimization.

\section{Related Work in \gls{fbo}}
\label{sec:related_work}

\gls{bo} is classically deployed in low dimensional settings ($P\approx 20$).
However, recently methods have been proposed which aim to make \gls{bo} more amenable to higher dimensional problems.
These include using random embeddings \citep{wang2016rembo,letham2020alebo,nayebi2019hesbo}, additive structure \citep{kandasamy2015additive,rolland2018additive,han2021tree}, and
variable selection \citep{papenmeier2022baxus,eriksson2021saasbo}.
Recent works have also re-examined the amenability of simple \gls{bo} strategies to scale to high dimensions \citep{hvarfner2024vanilla,doumont2025we}.
In this section, we rather review the established literature on the $P=\infty$ case.
As mentioned in the introduction, the overwhelming majority of optimization studies using \gls{bo} to optimize over functions first assume a finite dimensional parameterization before proceeding with standard methods.
Here we provide an overview of studies which have instead attempted to treat the function space directly, and which will serve as baselines in the numerical experiments.

\paragraph{Explicit Subspace Methods}
\cite{kundu2025active} compare two methods which operate by searching a $K<\infty$ dimensional linear subspace $\mathcal{S}\subset\Hs$ generated by the span of given functions $s_1, \ldots, s_K$.
The first generates these functions via a random sampling mechanism on $\Hs$, and the second uses the leading eigenfunctions of an integral operator defined with reference to the objective $f$.
However, the latter procedure assumes that functional gradients of $f$ can be evaluated, which is not available in our black-box setting.
We therefore focus on exploring the span of random functions.

\cite{shilton2020sequential} propose a more sophisticated method which searches within a sequence of affine subspaces.
Given a current optimal function $\vu_0$, it generates a random set of functions $s_1, \ldots, s_K$ and searches within the $K$ dimensional space $\vu_0 + \textrm{span}(\{s_1, \ldots, s_K\})$ for a fraction $\rho B$ of the overall budget $B$.
Then, it resamples a new set $s_1, \ldots, s_K$ and searches within the span of those new points offset by the new optimal $\vu$.
This avoids choosing a single finite dimensional subspace to search within, and operates somewhat like an infinite dimensional analog of \cite{kirschner2019adaptive}.
\cite{vellanki2019bayesian} take a parametric approach, using Bernstein polynomials of degree $k$ as candidate functions.
The Bernstein basis allows for the easy enforcement of shape priors, such as monotonicity.
However, the Bernstein parameterization is most natural in one dimension, and it is unclear how to extend the approach to higher dimensional $\D$.

\paragraph{Direct Optimization in the Function Space}

\cite{vien2018bayesian} showed that, when EI or UCB is maximized by gradient descent in $\Hs$ starting from a candidate $\vu_0$ with finite $L^0$ norm, every iterate remains in the span of the support functions of $\vu_0$ and of the $N$ previously observed functions $\vu_1,\ldots,\vu_N$ (\cref{eq:vien_support}), so that only the coefficients of those support functions evolve.
This is a positive result in a computational sense: it could have been the case that a single step of gradient descent yields a function which does not have a finite $L^0$ norm, and thus cannot be represented on a computer.
Luckily, however, we need only update coefficients.

However, this is a double-edged sword: the support locations are never optimized, and are instead fixed by the random initialization $\vu_0$ of each acquisition search and by the past observations.
Expressivity must therefore come from the number of support functions: the random initializations must have a high $L^0$ norm, denoted $v$, and every candidate additionally inherits the supports of all past observations.
As the number of observations $N$ grows, the complexity of computing the \gls{gp} kernel matrix grows with $v^2N^2$.
\cite{vien2018bayesian} find that, to manage computation, it is therefore necessary to conduct a post-hoc sparsification step after optimizing the acquisition function.
This involved deploying \gls{kmp} \citep{vincent2002kernel} to reduce to only $m\ll v$ many support functions.
Then, kernel computation is only of complexity 
$v^2N_{\textrm{init}}^2 + v m N_{\textrm{init}}N + m^2N^2 = O(m^2N^2)$, where $N_{\textrm{init}}$ is the size of the initial design.

There are two major limitations to this approach.
First, even if the initial points have a high $L^0$ norm, there is no guarantee that they are suitable for approximately optimizing $f$, especially if $\D$ is of moderate dimension.
Though motivated as performing optimization directly in $\Hs$, each acquisition of \cite{vien2018bayesian} is ultimately restricted to a finite dimensional linear subspace of $\Hs$, spanned by support functions that are drawn at random or inherited from past observations, rather than chosen to improve the acquisition. 
Second, the post-hoc sparsification step creates a critical issue: the sparse approximation of the optimum is not necessarily the optimum among sparse candidates.
This situation is similar to solving an integer program by first solving a continuous relaxation and then rounding. 
In the next section, we will propose a method which addresses both of these limitations.

\section{Methodology}
\label{sec:methodology}


We propose to directly optimize over the set of functions we can feasibly represent, i.e. the ones with a sparse representation in $\F$, say of size $m$. 
However, we maintain and even improve expressibility by allowing the support functions themselves to be selected, not only their coefficients.
This means jointly choosing $\alpha_1,\ldots,\alpha_m\in\mathcal{A}$ for a suitable coefficient region $\mathcal{A}$, as previous methods have, but also $x_1,\ldots,x_m\in\D$ to define a candidate function $\vu = \sum_{i=1}^m \alpha_i k(\cdot, x_i)$.
For a given $m$, we therefore search the set
$\mathcal{M}^m = \{u \in \Hs : \Vert u \Vert_0 \leq m\}$.
This set is not a linear subspace: for instance, $\kernelH(\cdot, x_1)$ and $\kernelH(\cdot, x_2)$ both belong to $\mathcal{M}^1$, but $\kernelH(\cdot, x_1) + \kernelH(\cdot, x_2)$ does not in general.
Nor is it a single manifold, but rather a stratified set $\mathcal{M}^m = \bigcup_{j\leq m} S^j$, whose strata $S^j = \{\vu\in\Hs : \Vert \vu \Vert_0 = j\}$ collect the functions of exact sparsity $j$.
For $\D\subseteq\R^d$ and the \gls{rbf} kernel, the canonical parameterization $\phi(\alpha, x) = \sum_{i=1}^m \alpha_i\kernelH(\cdot,x_i)$ maps the open set $W^m = \{(\alpha,x) : x_i \text{ pairwise distinct},\ \alpha_i\neq 0\}\subset\R^{(d+1)m}$ onto the top stratum $S^m$ as an $m!$-sheeted covering, the sheets differing only by a permutation of the support points.
The top stratum is therefore an immersed smooth manifold of dimension $l:=(d+1)m$, while the lower strata $S^j$, $j<m$, are reached at the boundary of $W^m$, where a coefficient vanishes or two support points coincide.
The sets are nested such that $\mathcal{M}^1\subset\mathcal{M}^2\subset\ldots$, and their union is the pre-\gls{rkhs} $\Hs_0$.
For general $\D$, there are functions in $\Hs$ that do not belong to any $\mathcal{M}^m$.
However, since $\Hs_0$ is dense in $\Hs$, any function in $\Hs$ can be approximated arbitrarily well by a member of some $\mathcal{M}^m$, and the optimal value of a continuous objective over $\mathcal{M}^m$ converges to its infimum over $\Hs$ as $m$ grows.
For moderate $m$, we therefore expect this method to perform well when the functional $f$ can be approximately minimized using functions with small $L^0$ norm.
From the function space perspective, at each iteration we solve:
\begin{align}
    \max_{u\in\Hs} \quad \acq(u) \quad s.t. \quad\,\,\Vert u \Vert_0 \leq m \,,
\end{align}
which practically speaking is implemented as:
\begin{align}
    &\max_{\substack{x_1, \ldots, x_m\in\D 
    \\\alpha_1, \ldots, \alpha_m\in\mathcal{A}}} 
    \quad 
    \acq\left(\sum_{i=1}^m
    \alpha_i k(\cdot,x_i)
    \right)  \,.
    \label{eq:acq_problem}
\end{align}
This involves solving a finite dimensional search to which we can apply the standard tools used to solve the \gls{bo} acquisition subproblem.
Over all of $\Hs$ the maximum need not exist, since sparse sets are not compact and the acquisition need not be continuous in their topology.
In \cref{eq:acq_problem}, however, the search is over a compact parameter set (a compact $\D$ and a compact $\mathcal{A}$, e.g.\ the box constraints of the bounded parameterization below) through the continuous map $\phi$, and the \gls{gp} posterior mean and standard deviation, and hence \gls{ei}, are continuous in the \gls{rkhs} distance for the kernels we use, so that a maximizer exists by the extreme value theorem.


\paragraph{Natural Gradient on the Manifold.}
A subtlety arises when one optimizes over $\mathcal{M}^m$ via a parameterization $\phi : \R^l \to \Hs$. The geometry that matters for the acquisition functional lives in $\Hs$, but a naive optimizer in the parameter space $\R^l$ uses the Euclidean metric, which is in general unrelated to $\Hs$-distance.
Equal Euclidean steps in $\R^l$ correspond to unequal $\Hs$-displacements, so a standard Euclidean gradient takes a step that is not aligned with the steepest direction on the manifold.
The classical remedy is the natural gradient \citep{amari1998natural}. Recall that gradient descent can be derived as the repeated solution of quadratic problems of the form:
\begin{equation}
    w_{t+1} = \underset{w}{\operatorname{argmin}}\; f(w_t) + \nabla f(w_t)\transposed (w - w_t) + \frac{1}{2\lambda}\norm{w - w_t}_2^2,
\end{equation}
which yields the update $w_{t+1} = w_t - \lambda \nabla f(w_t)$. Replacing the squared Euclidean distance with a non-Euclidean dissimilarity $D(w, w_t)^2$ produces the natural gradient update
\begin{equation}
    w_{t+1} = w_t - \lambda G^{+} \nabla f(w_t),
\end{equation}
where $G$ is the metric tensor associated with $D$ and $G^{+}$ its Moore--Penrose pseudoinverse.\footnote{On the top stratum $W^m$ the metric tensor of the canonical parameterization is positive definite and $G^{+}=G^{-1}$. It degenerates at the boundary, where the parameterization stops being an immersion: if $\alpha_i = 0$, for example, any displacement of $x_i$ leaves $\phi$ unchanged. The pseudoinverse discards these directions (see \cref{app:natural_gradient} for a truncated variant).}

For a smooth parameterization $\phi: \R^l \to \Hs$, the natural choice of $D$ is the $\Hs$-distance pulled back through $\phi$: $D(w, w') = \norm{\phi(w) - \phi(w')}_\Hs$. The resulting metric is the \emph{pullback metric} with metric tensor
\begin{equation}
    G_{ij}(w) = \inner{\partial_i \phi_w}{\partial_j \phi_w}_\Hs,
    \label{eq:pullback}
\end{equation}
which can be expressed as the cross-partial derivative of the \gls{rkhs} inner product:
\begin{equation}
    G_{ij}(w) = \partialover{v_i}\partialover{v'_j}\inner{\phi(v)}{\phi(v')}_\Hs \bigg|_{v=v'=w}.
    \label{eq:pullback_autodiff}
\end{equation}
This formula reduces computation of the metric tensor to differentiating a scalar function of $v$ and $v'$, which can be handled by any autodifferentiation framework, regardless of the choice of parameterization.

\paragraph{Canonical Parameterization.}
The simplest concrete choice is the canonical representation with a fixed number $m$ of support points:
\begin{equation}
    \phi(w) = \sum_{i=1}^m \alpha_i \kernelH(\cdot, x_i), \qquad w = (\alpha, x) \in \R^m \times \R^{m \times d}.
\end{equation}
The \gls{rkhs} inner product between two such functions is
\begin{equation}
    \inner{\phi(v)}{\phi(v')}_\Hs = \alpha\transposed K(x, x') \alpha',
\end{equation}
where $K(x, x')_{ij} = \kernelH(x_i, x'_j)$. The pullback metric tensor has block structure
$
    G = \begin{pmatrix} G_{\alpha\alpha} & G_{\alpha x} \\ G_{x\alpha} & G_{xx} \end{pmatrix},
$
where each block is obtained by differentiating $\alpha\transposed K(x, x') \alpha'$ in $(\alpha, x)$ and $(\alpha', x')$ . 

Crucially, this parameterization keeps the candidate $\phi(w)$ sparse with exactly $m$ support points at all times, and allows joint optimization over both coefficients and locations.

\paragraph{Bounded Parameterization.}
If additional properties are desired, such as boundedness of the candidate function, an alternative parameterization can be used. Instead of parameterizing directly by $(\alpha, x)$, we use an implicit parameterization $(y, x)$ where $y \in [y_{\min}, y_{\max}]^m$ are the (approximate) function values at the support points. The coefficients are derived via
\begin{equation}
    \alpha = (K_{xx} + \varepsilon I)^{-1} y.
\end{equation}
This gives $\phi_w(x) = K_{xx}\alpha = K_{xx}(K_{xx} + \varepsilon I)^{-1} y \approx y$ when the support points are sufficiently far apart, while the $\varepsilon$-regularization shifts all eigenvalues of $K_{xx}$ by $\varepsilon$ to ensure that $\alpha$ is well defined even when $K_{xx}$ is near-singular. Boundedness of $\phi(w)$ at the support points is enforced by constraining $y \in [y_{\min}, y_{\max}]^m$. The pullback metric for this parameterization follows from \cref{eq:pullback_autodiff}.

\paragraph{Quasi-Newton Optimization.}
Although natural gradient descent provides a geometrically motivated update, the standard practice in \gls{bo} is to use a quasi-Newton method such as L-BFGS-B \citep{byrd1995lbfgsb}, which builds an empirical preconditioner on top of the one provided by the natural gradient. We found this to give strong empirical performance on the resulting finite-dimensional problem (see \cref{app:natural_gradient}). The bounded parameterization enforces box constraints on $y$, and modern autodifferentiation frameworks compute the gradient with respect to $(y, x)$ at negligible cost. We use multi-start L-BFGS-B in all experiments.

\paragraph{Recovering the Updates of \cite{vien2018bayesian}.}
The proposed framework includes the gradient descent procedure of \cite{vien2018bayesian} as a special case. They show that for a surrogate \gls{gp} with squared exponential kernel $\kernelGP(\vu, \vu') = e^{-\frac{1}{2\ell}\norm{\vu-\vu'}_\Hs^2}$, the Fr\'echet derivative of $\kernelGP$ with respect to $\vu$ is the linear functional
\begin{equation}
    D_\vu \kernelGP(\vu, \vu') = -\frac{\kernelGP(\vu, \vu')}{\ell}\inner{\cdot}{\vu - \vu'}_\Hs,
\end{equation}
whose Riesz representer in $\Hs$ is $-\frac{\kernelGP(\vu, \vu')}{\ell}(\vu - \vu')$. For $\vu = \sum_{i=1}^{m_\vu} \alpha_i \kernelH(\cdot, x_i)$ and $\vu' = \sum_{i=1}^{m_{\vu'}} \alpha'_i \kernelH(\cdot, x'_i)$, this representer lies in $\Hs$ with support points given by the union of those of $\vu$ and $\vu'$. Since the gradient of the acquisition involves $\kernelGP(\vu, \vu_n)$ for every observed function $\vu_n$, $n=1,\ldots,N$, after $t$ steps of gradient descent from $\vu_0$ the candidate $\vu_t$ accumulates the support points of all observed functions:
\begin{equation}
    \vu_t = \sum_{i=1}^{m_{\vu_0}} \beta_i \kernelH(\cdot, x_i) + \sum_{n=1}^{N}\sum_{i=1}^{m_{\vu_n}} \beta_{n,i} \kernelH(\cdot, x_{n,i}),
    \label{eq:vien_support}
\end{equation}
where $x_{n,i}$ are the support points of $\vu_n$.
To interpret this as natural gradient descent, fix the support set $\{x_i\}_{i=1}^m$ to the union of the supports of $\vu_0$ and of all observed functions, and consider the parameterization:
\begin{equation}
    \phi(w) = \sum_{i=1}^m w_i \kernelH(\cdot, x_i), \qquad w \in \R^m.
\end{equation}
The pullback metric tensor of \cref{eq:pullback} for this parameterization is:
\begin{equation}
    G_{ij} = \inner{\partial_i \phi}{\partial_j \phi}_\Hs = \inner{\kernelH(\cdot, x_i)}{\kernelH(\cdot, x_j)}_\Hs = \kernelH(x_i, x_j),
\end{equation}
i.e.\ $G$ is the kernel matrix of $\kernelH$ evaluated on the fixed support set. Substituting $\vu = \phi(w)$ and $\vu' = \phi(w')$, the squared norm reduces to $\norm{\phi(w) - \phi(w')}_\Hs^2 = (w - w')^\top G (w - w')$, and the Euclidean gradient with respect to $w$ is:
\begin{equation}
    \nabla_w \kernelGP(\phi(w), \phi(w')) = -\frac{\kernelGP(\phi(w), \phi(w'))}{\ell} \, G(w - w').
\end{equation}
The update direction of \cite{vien2018bayesian}, expressed as a coefficient vector in this parameterization, is instead $-\frac{\kernelGP(\phi(w), \phi(w'))}{\ell}(w - w')$, which is exactly the Euclidean gradient premultiplied by $G^{-1}$ (or $G^{+}$ if support points repeat). The updates of \cite{vien2018bayesian} are therefore equivalent to natural gradient descent in this parameterization.

\section{Experimental Results}
\label{sec:experiments}

We compare \wycoff{} against \vien, \vellanky, \kundu, and \shilton, described in \cref{sec:related_work}, on function projection problems, on ridge functionals derived from classical \gls{bo} benchmarks, and on real-world tasks including control problems and functional hyperparameter optimization.
All objectives are minimized and their optimal value is not necessarily zero.

We investigate two metrics of interest: the best target value and average regret.
For each metric, we report the median and 95\% \gls{ci} across independent runs.
All methods share the same budget and complexity schedule: after 10 initial designs, the size $k$ of the acquisition problem grows from $1$ to $10$, increasing every $10$ acquisitions, for a total of $110$ evaluations (\cref{app:implementation}); \cref{app:fixed_k} compares this schedule with a fixed $k$.
The optimal \gls{rkhs} lengthscale and the \gls{gp} kernel are selected for each method and benchmark (\cref{app:kernel_profile,app:ablations}).
As a baseline, we report the value obtained by following the same schedule without optimizing the acquisitions (\random), i.e.\ sampling from the \gls{rkhs} at random.
A breakdown of wall-clock times is given in \cref{app:timing}.

\paragraph{Projections.}
As a simple toy problem, we project a fixed target function $\tilde{f}$ onto the \gls{rkhs}, minimizing the squared $L^2$ distance between the candidate and the target.
We use the \emph{sinc} function, scaled and centered so that its main oscillations lie within the unit domain $[0,1]^d$:
\begin{equation}
    \tilde{f}(x) = \frac{1}{d}\sum_{i=1}^d \operatorname{sinc}\!\left(2 \pi x_i - \pi\right);
    \quad
    f(\vu) = \Vert \tilde{f} - \vu\Vert_{L^2}^2 \,,
\end{equation}
where the integral is computed with a midpoint quadrature on a tensor grid of about $4096$ points.
We consider input dimensions $d \in \{1, 2, 3, 4, 5\}$ to evaluate how the methods perform on a similar task at varying dimensionalities.

\paragraph{Ridge Functions.}
A large corpus of benchmark functions exists for standard \gls{bo}, but comparatively few for \gls{fbo}.
We propose a systematic approach to adapt such benchmarks to the functional setting, extending the concept of \emph{ridge functions} to the functional domain.
Given a scalar benchmark $\bar{f}:[0,1]^d\to\R$ and a candidate function $\vu: [0,1]^d \to \R$, we define the functional benchmark as
\begin{equation}
    f(\vu) = \bar{f}\!\left(\sigma\big(b + L[\vu]\big)\right), \qquad L_i[\vu] = \int_{[0,1]^d} w_i(x)\, \vu(x)\, dx, \quad i=1,\ldots,d,
\end{equation}
where $\sigma$ is the logistic function applied elementwise, $b\sim\mathcal{U}(-1,1)^d$ is a random bias, and each weight $w_i$ is a random trigonometric polynomial with frequencies $\Vert \omega \Vert_1 \leq 4$ and decaying amplitudes, normalized to $\Vert w_i \Vert_{L^2}=1$.
The functionals $L_i$ are not represented by a sparse element of $\Hs$, so we do not anticipate this construction to unduly favor our method.
For \gls{rbf} \gls{rkhs} candidates the integrals factorize across dimensions and are computed in closed form via the complex error function; for other candidate representations we use a midpoint quadrature.
Evaluating $L_i$ by averaging $w_i \vu$ at $n$ random points instead yields a Monte Carlo estimate with $O(1/\sqrt{n})$ error, and we verify this convergence in \cref{app:convergence}.

We provide a \textsc{jax}~\citep{jax2018github} compatible open-source implementation of the virtual library of \cite{surjanovic2013virtual}, comprising $48+$ benchmark functions, together with the functional ridge extension.
In this work we consider five popular benchmarks of increasing input dimensionality: \emph{GramacyLee} ($d=1$), \emph{Ackley} ($d=2$), \emph{Hartmann} ($d=3$), \emph{Rosenbrock} ($d=4$), and \emph{Michalewicz} ($d=5$).

\paragraph{Real World Tasks.}
We next consider problems where the functional parameter has a direct physical or model interpretation.
The \emph{Brachistochrone} problem asks for the curve $y:[0,1]\to\R$ passing through two given points such that a bead placed at the start point and sliding along it under gravity reaches the end in the minimum time.
The optimal solution is known to be a cycloid and the optimal travel time is known.
Efforts to analytically solve the Brachistochrone problem precipitated the development of variational analysis, and we think this makes the Brachistochrone a suitable testbed for \gls{fbo}.
We optimize for the travel time in excess of that of the cycloid.

The second benchmark is the optimization of the activation function of a neural network. A better activation can speed up convergence and increase accuracy. We train a simple \gls{mlp} with two hidden layers of 256 neurons on the MNIST digit recognition task \citep{lecun1998mnist}. Our setup differs slightly from \cite{vien2018bayesian}: we measure the average test error across multiple training runs rather than a single seed, and instead of optimizing for the activation directly, we optimize an offset relative to the popular ReLU activation.

\begin{figure}[h]
    \centering
    \includegraphics[width=0.4\textwidth]{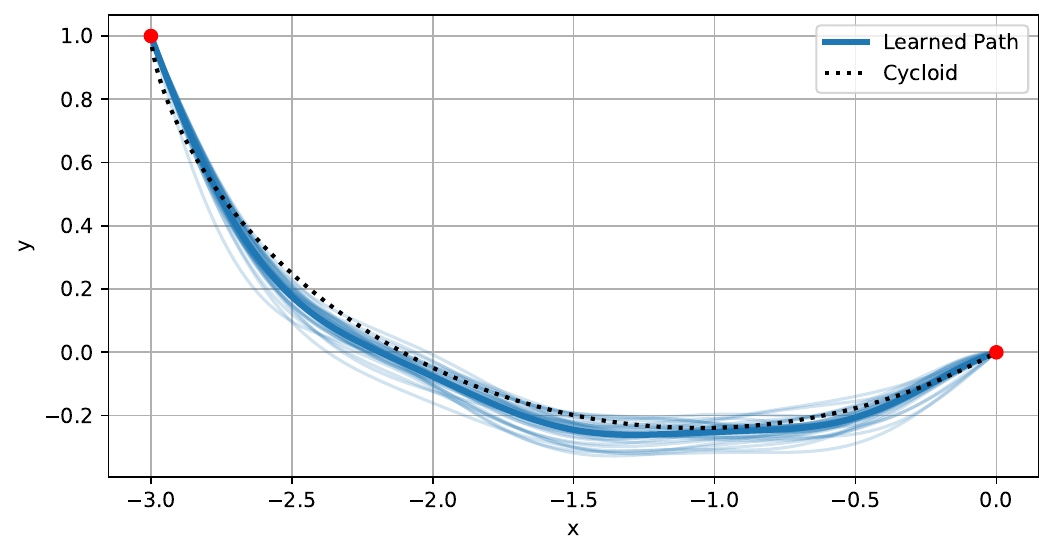}
    \hspace{0.04\textwidth}
    \includegraphics[width=0.4\textwidth]{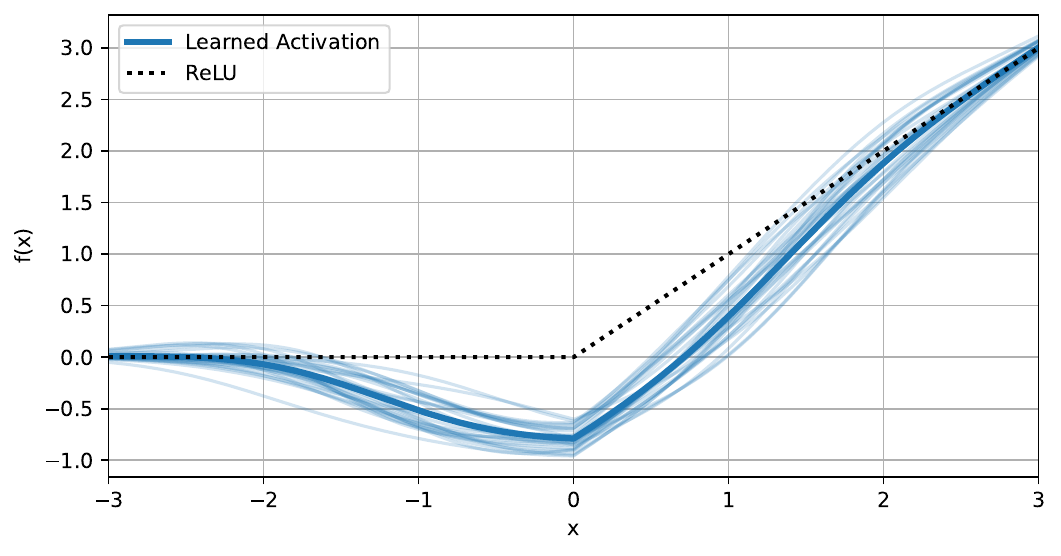}
    \caption{Best functions found by \wycoff{}: brachistochrone paths (left) and MNIST activation functions, i.e.\ ReLU plus the learned offset (right). Transparent lines show individual runs; bold lines show the average across runs.}
    \label{fig:f_vis}
    \vspace{-0.1in}
\end{figure}

\paragraph{Robotics Control Tasks.}
We also consider three control tasks where the policy itself is the functional parameter being optimized. The first is the classic underactuated pendulum environment from OpenAI Gym \citep{brockman2016openai}, where the objective is to swing up and balance the pendulum by applying a torque $\vu:\R^2\to\R$ as a function of the current angle and angular velocity. The objective is the expected cumulative reward over an episode, evaluated over 50 rollouts.
This problem is more challenging than the typical \gls{rl} setting since we are optimizing the policy without access to state transitions; gradient-based policy optimization cannot be applied.

The second is a more complex environment in which a double-jointed robot arm must interact with a pinwheel and rotate it to a given target angle. The policy is a zero-force (i.e., desired) trajectory of the shoulder joint $\vu:[0,1]\to\mathbb{R}$ for a joint-space impedance controller \cite{hermus2021exploiting}. The joint stiffness and damping parameters in the controller are fixed, as well as the zero-force equilibrium point of the elbow joint. 
The objective is the cosine distance between the final pinwheel angle and the desired one.
The arm must make and break contact with the pinwheel multiple times while respecting its kinematics; such hybrid problems, with both continuous motions and a discrete contact sequence, are challenging for standard solvers due to the discontinuity of the gradient at the contact transitions~\cite{posa2014direct}.

The third, \emph{Hopper}, is a planar single-leg robot (body, thigh, shank and foot) with rigid ground contact and Coulomb friction, which must jump as high as possible. The policy is a zero-force (i.e., desired) trajectory of the hip and knee joints $\vu:[0,1]\to\R^2$ over the $2$\,s episode for a joint-space impedance controller. The joint stiffness and damping parameters in the controller are fixed. The ankle was held at a fixed angle.
The objective is the negative maximum height of the body after the first landing, in metres.
As in the pinwheel task, intermittent contact makes the objective discontinuous in the policy.

\begin{figure}[h]
    \centering
    \includegraphics[width=\textwidth]{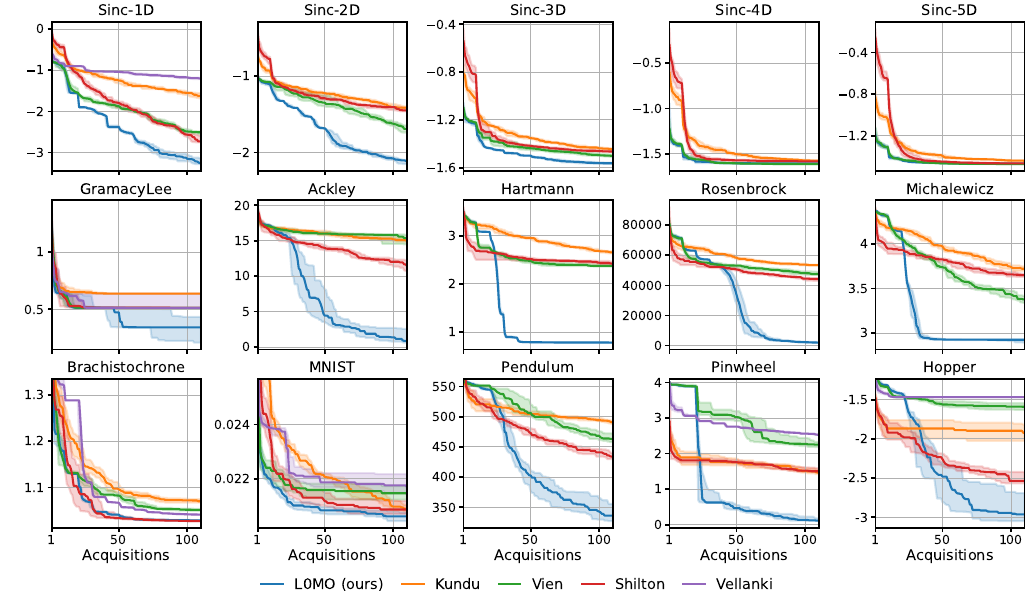}
    \caption{Method comparison across benchmark functions: projections (top), functional ridges (middle) and real-world tasks (bottom). Each plot shows the running best target function value (median and 95\% \gls{ci}) as a function of the number of acquisitions.}
    \label{fig:method_comparison}
    \vspace{-0.1in}
\end{figure}

\paragraph{Results.}
\Cref{fig:method_comparison} and \cref{tab:regret_combined} report the results.
On the projections, \wycoff{} attains the lowest median average and final regret on Sinc-1D, Sinc-2D and Sinc-3D, and the lowest or tied median average regret on Sinc-4D and Sinc-5D, where \vien{} has a marginally lower final value.
\wycoff{} also attains the lowest median average and final regret on all five ridge functionals, with the largest margins on Ackley, Hartmann and Rosenbrock, where the competitors improve only moderately over random sampling.
On the real-world tasks the gaps are smaller.
\wycoff{} has the lowest median on Pendulum, Pinwheel and Hopper, and also on MNIST, although there all methods lie within about a tenth of a percentage point of test error and the \glspl{ci} of \wycoff, \kundu{} and \shilton{} overlap.
On Brachistochrone, \shilton{} reaches a slightly lower final regret, although \wycoff{} has the lower median average regret.
We attribute this to \shilton's search being centered at the incumbent and spanned by smooth \gls{gp} prior samples, which favors the local refinement of a smooth, non-sparse optimum such as the cycloid late in the run.
\Cref{fig:f_vis} shows the functions found by \wycoff{} on Brachistochrone and MNIST.

\input{tables/results_combined}

\section{Discussion, Conclusion and Future Work}
\label{sec:discussion}

We proposed a new method for \gls{fbo} which optimizes over a sparse subset of an \gls{rkhs}, jointly searching over basis point locations and coefficients. We showed that the gradient descent procedure of \cite{vien2018bayesian} is equivalent to natural gradient descent in a finite-dimensional parameter space, and that this perspective generalizes to any smooth parameterization. We deployed our method on a range of synthetic and real-world benchmarks, where it attained the best or tied-best median performance on most of them.

\paragraph{Limitations.}
Our analysis is not a convergence or regret guarantee: density of the sparse sets only ensures that good sparse approximations exist as $m$ grows, and the acquisition landscape over the joint $(\alpha, x)$ space is highly non-convex, so multi-start L-BFGS-B, although effective in practice, may not find its global optimum.
The method is best suited to objectives that can be approximately minimized by functions with few support points, and a smooth, non-sparse optimum such as the cycloid may be better served by dense bases, as the Brachistochrone results suggest.
Our experiments use a complexity schedule shared across all methods, which some methods may not benefit from equally (\cref{app:fixed_k}), and a relatively coarse grid of \gls{rkhs} lengthscales and \gls{gp} kernels, so a finer search might reveal stronger configurations for one or more comparators.
Finally, the cost of each acquisition grows with the number of support points and observations, which, as for any \gls{gp}-based method, limits the approach to moderate evaluation budgets (\cref{app:timing}).

We see several promising directions for future work. First, the complexity $m$ is currently set by a fixed schedule; an adaptive schedule which selects $m$ based on the data would be a natural extension. Second, our framework treats each acquisition independently; warm-starting the optimization from previously good candidates may further improve sample efficiency. Third, the natural gradient perspective opens up the use of other parameterizations, including those which enforce structural priors such as monotonicity or periodicity. Finally, the results on control problems suggest that \gls{fbo} may be a viable strategy for policy optimization in low-data regimes; a more thorough study of this application is warranted.

\bibliographystyle{plain}
\bibliography{references}
\newpage


\appendix
\FloatBarrier
\section{\gls{rkhs} Kernel Scale and Continuity Priors}
\label{app:ablations}

Following the previous literature, we use an \gls{rkhs} with an \gls{rbf} kernel.
The lengthscale parameter $\ell$ determines which functions admit a sparse approximation, and therefore changes the optimization landscape.
Since the optimal lengthscale is task-specific rather than method-specific, we consider a log-uniform grid of values $\{0.4, 0.2, 0.1, 0.05\}$, shared by all methods on a benchmark.
The selected lengthscale is the one at which the best configuration (method and \gls{gp} profile) among the five compared methods attains its lowest median final value; ablation variants of \wycoff{} do not take part in the selection.

The surrogate \gls{gp} kernel also plays a role.
The choice of kernel defines the prior on the smoothness of the surrogate model.
We consider four profiles: \gls{rbf}, Mat\'ern-5/2, Mat\'ern-3/2, and Mat\'ern-1/2, corresponding respectively to $\mathcal{C}^\infty$, $\mathcal{C}^2$, $\mathcal{C}^1$, and $\mathcal{C}^0$ surrogate models.
Unlike standard \gls{bo}, where the target function is often assumed to be at least differentiable, in the functional domain the target may be discontinuous, making rougher kernels potentially more appropriate.
The optimal profile may vary not only across target functions but also across methods, so we tune this independently for each method, showing each approach at its best.
The selected lengthscales and profiles for each setting are reported in \cref{tab:hyperparams}.

Since the lengthscale selection could favor one method, \cref{tab:lengthscale_ranks} reports the mean rank of each method over the ten main benchmarks at every lengthscale in the grid, and \cref{tab:lengthscale_full} reports the underlying medians.
\wycoff{} has the best mean rank for both metrics at every lengthscale, and its advantage is largest at the longer lengthscales, where the sparse candidates are smoother.

\input{tables/best_lengthscales}
\input{tables/lengthscale_ranks}
\input{tables/lengthscale_full}

\FloatBarrier
\section{Effect of Natural Gradient}
\label{app:natural_gradient}

\begin{figure}[h]
    \centering
    \includegraphics[width=\textwidth]{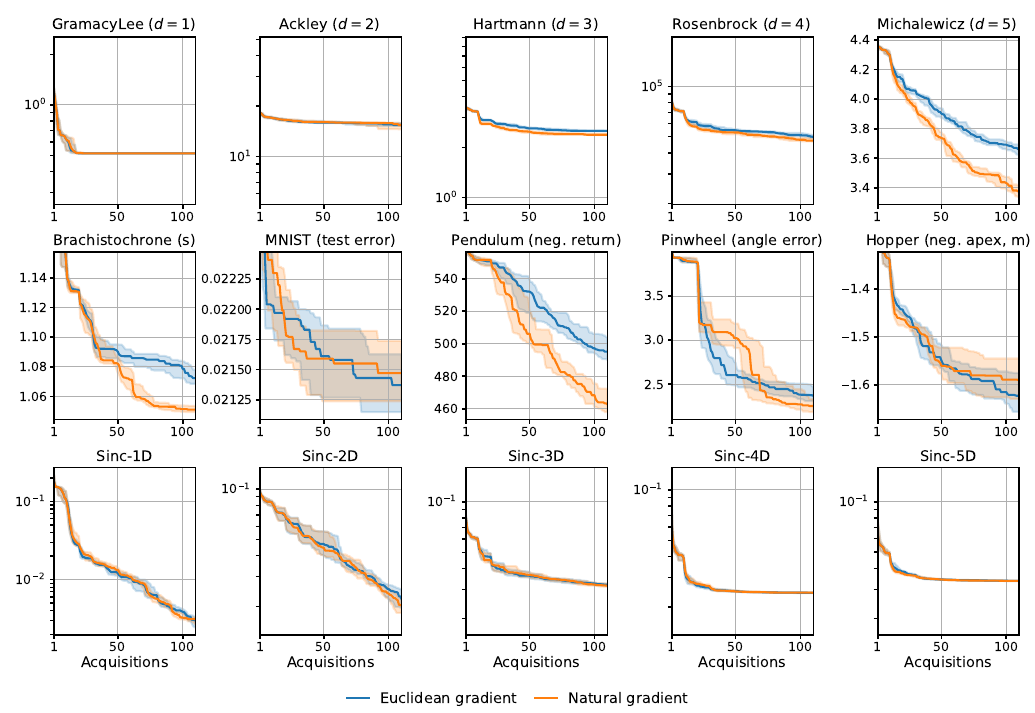}
    \caption{Effect of natural gradient preconditioning for \vien{} across benchmark functions.}
    \label{fig:gradient_ablation_vien}
\end{figure}

\Cref{fig:gradient_ablation_vien} compares the optimization trajectories obtained when running the procedure of \cite{vien2018bayesian} with and without the natural gradient preconditioner.
As established in \cref{sec:methodology}, the original updates of \cite{vien2018bayesian} correspond to natural gradient descent in a finite-dimensional parameter space; here we ablate the role of the preconditioner directly.
The preconditioner is the pseudo-inverse $G^{+}$ of the pullback metric, computed with the default singular value cutoff of the least squares solver.

The L-BFGS-B optimizer already builds an empirical approximation to the inverse Hessian from gradient history, which in practice subsumes much of the correction that the pullback metric would otherwise provide.
For \vien{}, the difference between running with and without the natural gradient preconditioner is small on most benchmarks, with a slight advantage for the natural gradient on Brachistochrone, Michalewicz and Pendulum.

For \wycoff{}, where the parameterization jointly optimizes over support locations $x$ and coefficients $\alpha$, applying the natural gradient preconditioner on top of L-BFGS-B markedly \emph{hurts} performance on GramacyLee, Ackley, Hartmann, Rosenbrock, Brachistochrone, Pendulum and Pinwheel, and slightly on Hopper; the two are comparable on the projection benchmarks and MNIST, and the natural gradient is better only on Michalewicz.
Unlike for \vien{}, whose support points are fixed during the acquisition optimization, the metric of \wycoff{} depends on the support locations and degenerates when two of them approach each other or a coefficient approaches zero, which makes the preconditioned updates poorly scaled near these points.
These results are consistent with quasi-Newton methods being effectively able to learn the proper metric such that explicitly using the preconditioner is superfluous.
Since the ablations in \cref{app:sampling_ablation,app:fixed_k} were run with the natural gradient variant of \wycoff{}, they should be compared with it rather than with the results of \cref{sec:experiments}.

\begin{figure}[h]
    \centering
    \includegraphics[width=\textwidth]{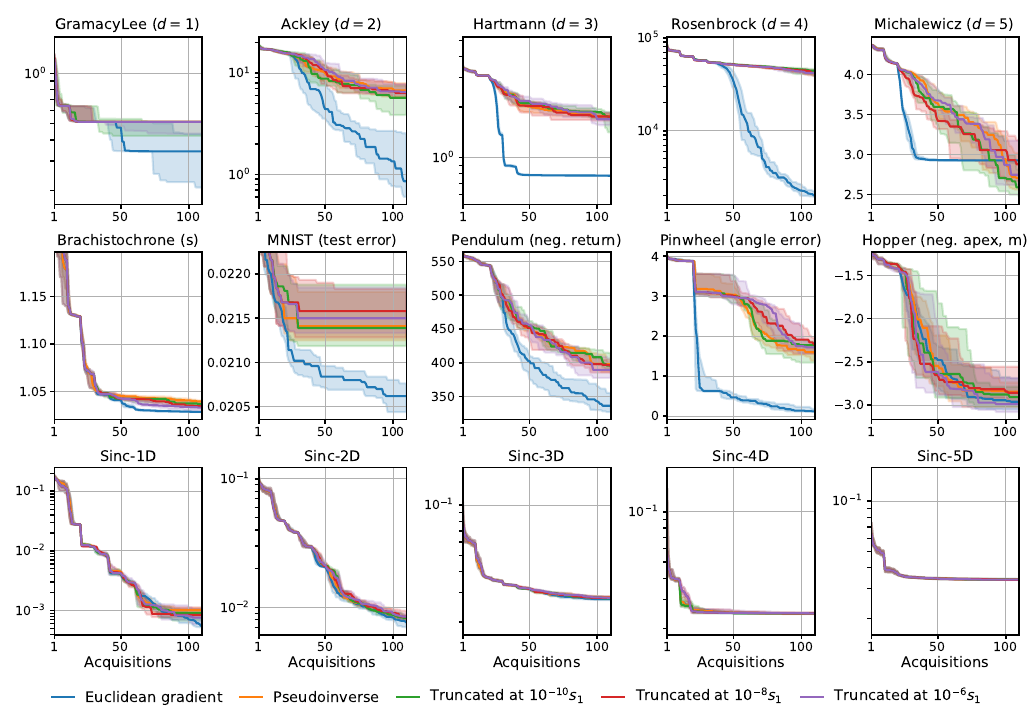}
    \caption{Effect of truncating the small singular values of the pullback metric in the natural gradient preconditioner of \wycoff{}, compared with the pseudo-inverse and the Euclidean gradient.}
    \label{fig:preconditioner_ablation}
\end{figure}

\paragraph{Regularization of the preconditioner.}
To test whether the degradation is caused by small singular values of $G$, \cref{fig:preconditioner_ablation} compares the pseudo-inverse with truncated pseudo-inverses that discard singular values $s_i < r\, s_1$ for $r \in \{10^{-10}, 10^{-8}, 10^{-6}\}$.
Truncation has no consistent effect beyond a small improvement on Brachistochrone: the variants remain close to each other, and all are worse than the Euclidean gradient on the benchmarks where the natural gradient hurts.

\begin{figure}[h]
    \centering
    \includegraphics[width=\textwidth]{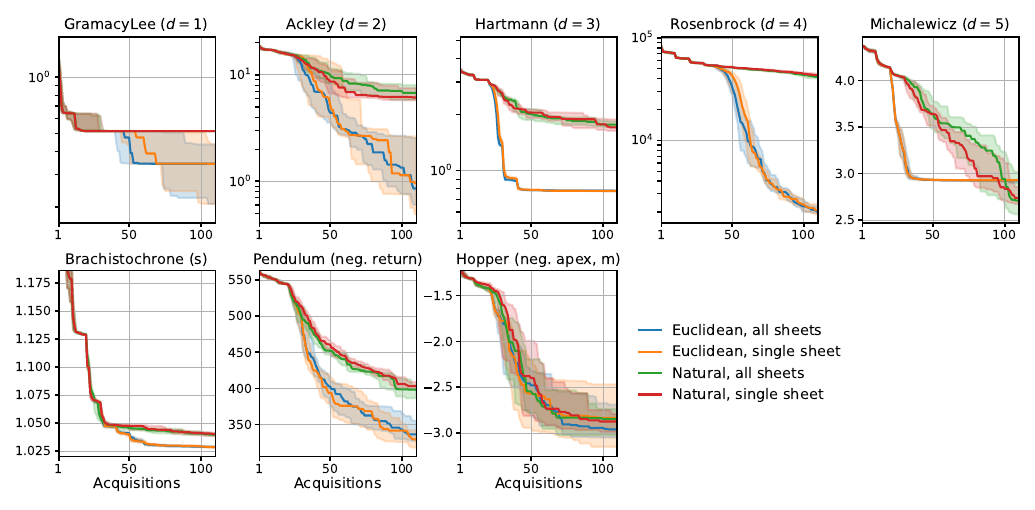}
    \caption{Effect of restricting \wycoff{} to a single sheet of the parameterization, with Euclidean and natural gradients, on the benchmarks where the single-sheet variant was run.}
    \label{fig:constrained_ablation}
\end{figure}

\paragraph{Symmetry of the parameterization.}
The parameterization $(\alpha, x)$ covers each sparse function $m!$ times, once per ordering of the support points (\cref{sec:methodology}).
To test whether this symmetry hinders the optimizer, we restrict the search to a single sheet by parameterizing the first input coordinate of the support points as the normalized cumulative sum of positive gaps, which sorts the support points along that coordinate.
\Cref{fig:constrained_ablation} shows that, on the ridge functionals and on Brachistochrone, Pendulum and Hopper, where it was run, the restriction has no appreciable effect with either gradient: the redundant sheets neither help nor hurt the multi-start optimization.

\FloatBarrier
\section{Complexity Schedule}
\label{app:fixed_k}

\begin{figure}[h]
    \centering
    \includegraphics[width=\textwidth]{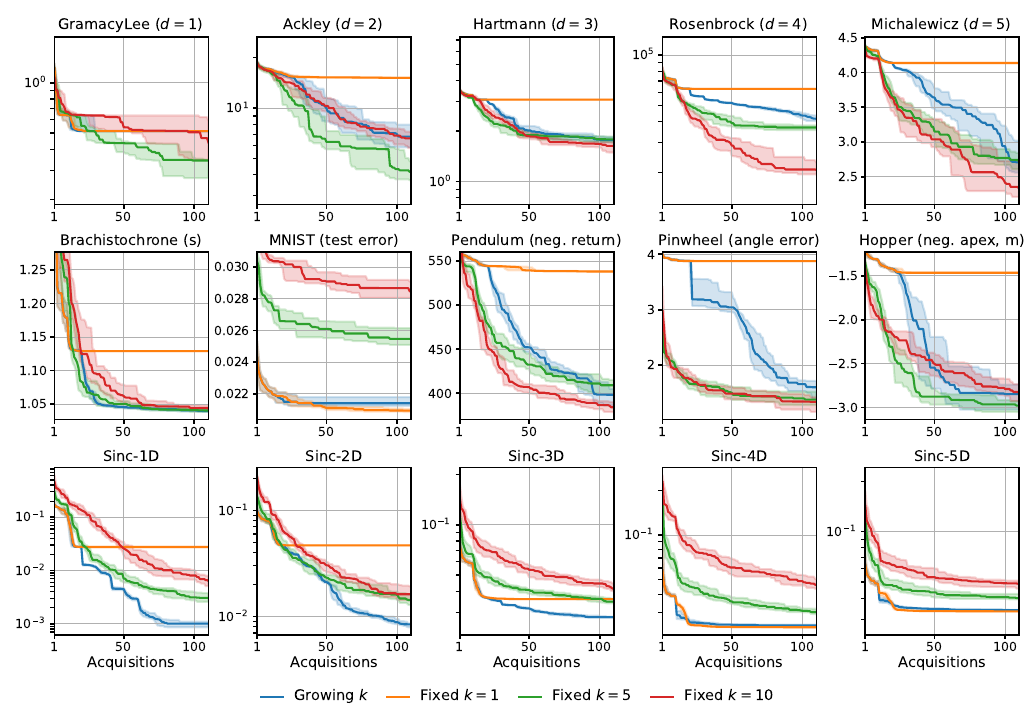}
    \caption{Growing complexity schedule compared with a fixed number of support points $k$ for \wycoff{}.}
    \label{fig:fixed_k_ablation}
\end{figure}

All methods in \cref{sec:experiments} share a schedule in which the complexity $k$ grows from $1$ to $10$ every $10$ acquisitions.
\Cref{fig:fixed_k_ablation} compares this schedule with keeping $k$ fixed at $1$, $5$ or $10$ throughout the run, for \wycoff{} (natural gradient variant).
A single support point ($k=1$) is the worst choice on nearly every benchmark, with the exceptions of MNIST, Sinc-4D and Sinc-5D, whose optima are well approximated by very simple functions.
Fixing $k=5$ or $k=10$ improves over the growing schedule on several ridge functionals and on the Pendulum, Pinwheel and Hopper tasks, but is worse on Brachistochrone and clearly worse on MNIST and the projection benchmarks.
No fixed complexity dominates, and the growing schedule is the most robust choice across benchmarks; an adaptive schedule that selects $k$ from the data is a natural direction for future work.

\FloatBarrier
\section{Gaussian Process Kernel Profile}
\label{app:kernel_profile}

The profile of the surrogate \gls{gp} kernel determines the smoothness assumptions placed on the target functional $f$.
In standard finite-dimensional \gls{bo}, the target is commonly assumed to be at least once differentiable, and the \gls{rbf} or Mat\'ern-5/2 kernel is a natural default.
In the functional domain, however, $f$ maps functions to scalars and need not inherit the regularity of the underlying function space; for instance, evaluation-based objectives such as travel time or cumulative reward can be discontinuous as a function of the \gls{rkhs} distance between candidates.
This motivates considering rougher kernels, including Mat\'ern-3/2 and Mat\'ern-1/2, as surrogate priors.

\begin{figure}[h]
    \centering
    \includegraphics[width=\textwidth]{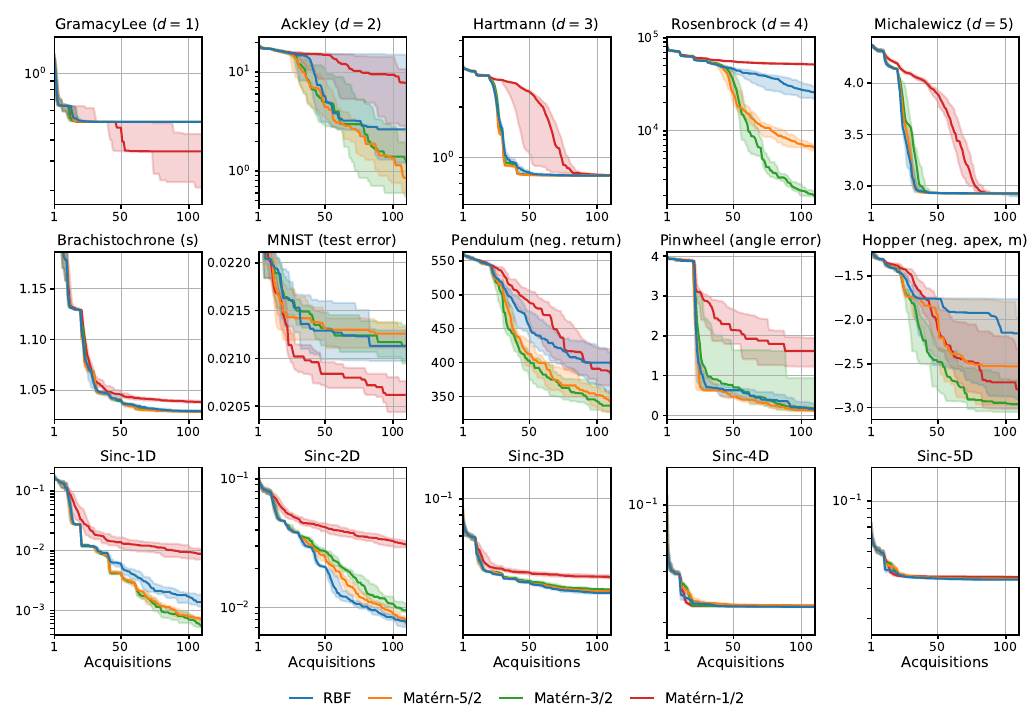}
    \caption{Effect of the \gls{gp} kernel profile on \wycoff{} across benchmark functions. Each line corresponds to a different smoothness assumption on the surrogate.}
    \label{fig:kernel_profile_ablation}
\end{figure}

\Cref{fig:kernel_profile_ablation} shows the performance of \wycoff{} under each of the four kernel profiles across all benchmarks, at the selected \gls{rkhs} lengthscale.
The profile matters substantially on Ackley, Rosenbrock, Pinwheel and Sinc-1D, where the median final value varies by an order of magnitude across profiles, and hardly at all on Hartmann, Michalewicz, MNIST, Sinc-4D and Sinc-5D.
Mat\'ern-5/2 and Mat\'ern-3/2 are the best or close to the best on most benchmarks, while the roughest profile, Mat\'ern-1/2, is the worst on most synthetic benchmarks and on Brachistochrone and Pinwheel, but the best on GramacyLee and MNIST.
No single profile dominates uniformly, and the selected profile also varies across methods on the same benchmark (\cref{tab:hyperparams}).
In the main experiments of \cref{sec:experiments}, each method is therefore evaluated at the profile that achieves the lowest median final value for that method and benchmark, so that no method is penalized by a suboptimal kernel choice.
\Cref{tab:matern52} shows that the conclusions of \cref{sec:experiments} do not depend on this per-method selection: with the Mat\'ern-5/2 profile for every method, \wycoff{} is tied for first on every benchmark and metric except the final value on Brachistochrone, where \shilton{} is marginally better.

\input{tables/matern52}

\FloatBarrier
\section{Initial Candidate Distribution}
\label{app:sampling_ablation}

At each acquisition step, the L-BFGS-B optimizer is initialized from a set of $16$ candidate starting points,
selected as the top $16$ out of $1024$ initial guesses evaluated on the acquisition function.
The default strategy samples these $1024$ guesses via \gls{lhs} in the parameter space, as described in \cref{app:implementation}.
Following \cite{shilton2020sequential}, an alternative is to draw the initial guesses from the surrogate \gls{gp} prior.

\begin{figure}[h]
    \centering
    \includegraphics[width=\textwidth]{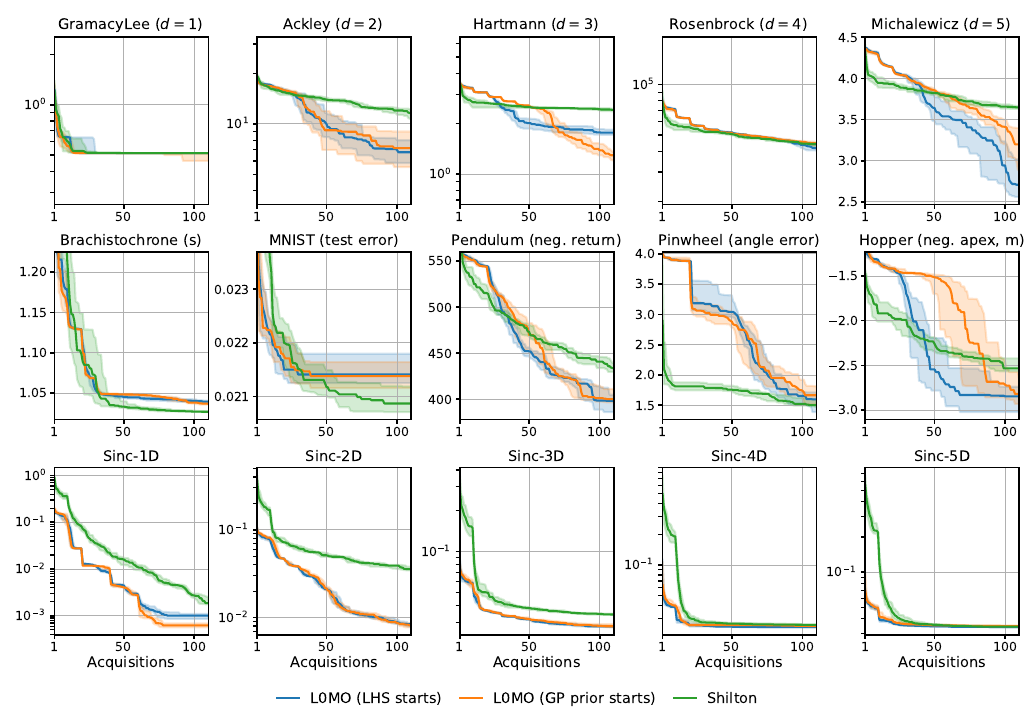}
    \caption{Effect of the initial candidate distribution across benchmark functions.}
    \label{fig:sampling_ablation}
\end{figure}

\Cref{fig:sampling_ablation} compares these two sampling strategies for \wycoff{} (natural gradient variant) across all benchmarks, overlayed with \shilton{} for reference.
In the early acquisitions, the choice of initial candidate distribution produces a visible difference: prior-based sampling can lead to better or worse starting points depending on how suitable the \gls{gp} prior is for the given test function.
As the number of acquisitions grows, the two strategies reach similar final values on most benchmarks, with differences of no consistent sign: prior-based starts are better on Hartmann and Brachistochrone, and worse on Michalewicz and, in average regret, on Hopper.
This is consistent with the expectation that, given a sufficiently good acquisition optimizer, the final solution found by L-BFGS-B is determined mostly by the acquisition landscape rather than by the initialization, and the top-$16$ selection already filters out poor starting points effectively.
These results suggest that the initial candidate distribution is not a critical design choice for \gls{fbo}.

\FloatBarrier
\section{Implementation Details}
\label{app:implementation}

The surrogate \gls{gp} is fit by maximum likelihood estimation of the lengthscale $\ell$ and nugget $g$, with the optimal trend $b$ and scale $\nu$ estimated analytically. For \wycoff, \vien, \kundu, and \shilton, the surrogate takes functional inputs $\vu$ with a stationary kernel defined on \gls{rkhs} distances $\|\vu - \vu'\|_\Hs$. For \vellanky, the surrogate works directly on the Bernstein polynomial coefficients, bypassing the \gls{rkhs} representation entirely.
Following \textsc{hetGP}~\citep{binois2021hetgp}, the lengthscale bounds are set from the $5\%$ and $95\%$ quantiles of the pairwise distances between observations, so that the closest pairs can decorrelate down to a correlation of $0.01$ and the farthest pairs can remain correlated at $0.5$; the nugget is bounded in $[\varepsilon_{\textrm{machine}}, 100]$.
The observations are zero-padded and masked to the next power of two, so that the likelihood and the predictive equations are compiled once per padded size rather than once per acquisition, and the LU factorization of the covariance matrix is computed once per fit and reused throughout the acquisition optimization.

The acquisition function used follows the numerically stable log-\gls{ei} implementation of \cite{ament2024logei}.
All methods use the same acquisition optimization strategy: given $1024$ initial candidates sampled via \gls{lhs}, the acquisition loss is evaluated at each candidate and the top $16$ are selected as starting points for L-BFGS-B optimization with \texttt{maxiter}$=100$, \texttt{ftol}$=\sqrt{\varepsilon_{\textrm{machine}}}$, and \texttt{gtol}$=0$. The best result across restarts is returned.

Since the surrogate \gls{gp} operates on \gls{rkhs} distances, it can be fit on functional inputs with varying numbers of support points. Therefore the complexity of candidate points can vary across acquisition steps.
Finding the optimal complexity schedule is beyond the scope of this work (see \cref{app:fixed_k} for a comparison with fixed complexities). For our evaluation we adopt a fixed schedule, shared across all methods. The parameter $k$ is taken to represent the size of the acquisition optimization problem: $k$ is the subspace dimension for \kundu{} and \shilton, the number of support points for \vien{} and \wycoff, and the polynomial degree for \vellanky.
We evaluate all methods on a fixed budget of $110$ function evaluations, beginning with 10 initial candidates sampled via \gls{lhs}, followed by 10 sequential acquisitions at each complexity level $k \in \{1, \ldots, 10\}$.
The \random{} baseline follows the same schedule but skips surrogate modeling, using \gls{lhs} for all acquisitions.

The \gls{rkhs} lengthscale is selected per benchmark from $\{0.4, 0.2, 0.1, 0.05\}$ and the \gls{gp} kernel profile per method and benchmark, as detailed in \cref{app:kernel_profile,app:ablations}.
Each configuration (benchmark, method, kernel profile and lengthscale) is repeated with 16 to 32 independent seeds, for a total of about $105$k individual runs.
The sweep was carried out on a Slurm cluster, and we provide a script to reproduce it, including preset time and memory requirements.
Each individual run can be reproduced on a consumer laptop, with running times ranging from a few minutes for the synthetic benchmarks to a few hours for neural network training and control simulations (\cref{app:timing}).

All benchmarks operate in the same function space, optimizing over candidate functions $\vu:[0,1]^d\to[-1,1]$.
To avoid discrepancies arising from differing input-output domains, additional care is needed to homogenize representation across methods.

\paragraph{\vien.}
Initial candidates are sampled via the distribution induced by \gls{lhs} and the bounded parameterization with $k_{\min}$ support points.
At each acquisition step, new candidate functions are first sampled via \gls{lhs} with $k$ support points.
Each candidate is then expanded to include the union of all historical observation support points $\{x_i\}$, with zero coefficients appended for the new points.
The function values at all expanded support points are soft-squished to $[-1,1]$ using a scaled logistic function.
Given this final step, the induced metric is no longer exactly given by the canonical \gls{rkhs} inner product. We compute it using the general pullback formula.
After optimization, \gls{kmp} \citep{vincent2002kernel} is applied to sparsify the result to exactly $k$ support points before target evaluation.

\paragraph{\kundu.}
The basis functions for the random subspace are sampled with $k_{\max}$ support points throughout.
At each acquisition step a fresh random subspace of dimension $k$ is constructed by sampling $k$ basis functions via the distribution induced by \gls{lhs} and the bounded parameterization.
The candidate is a linear combination of these basis functions with coefficients $c \in [-1,1]$, obtained by concatenating the support points and coefficients of all basis functions.
The corresponding $y$ values are computed and soft-squished to enforce the $[-1,1]$ range.
The final candidate is reconstructed using the squished values.

\paragraph{\shilton.}
At each acquisition step, a fresh random subspace of dimension $k$ is constructed by sampling $k$ basis functions from the surrogate \gls{gp} prior on a set of $n=50$ points in $[0,1]^d$ drawn via \gls{lhs}. The original paper suggests using $n=100$, but we found that denser grids led to numerical instability.
The subspace is centered at the incumbent, i.e.\ the observed function $\vu_n$ with the lowest target value $f(\vu_n)$ among $\vu_1, \ldots, \vu_N$.
The $y$ values of the linear combination at the support points are soft-squished to $[-1,1]$, and the candidate is reconstructed via the bounded parameterization and sparsified to $k_{\max}$ support points with \gls{kmp} before target evaluation.

\paragraph{\vellanky.}
Candidates are represented as Bernstein polynomials of degree $k$ with coefficients $c \in [-1,1]^{k+1}$, sampled via \gls{lhs} and rescaled to $[-1,1]^{k+1}$.
At evaluation time, the input is soft-squished to map to $[0,1]$ in order to make the method usable on the one-dimensional benchmarks.
When the complexity level $k$ increases, all existing observations are lifted to degree $k$ via the exact Bernstein degree elevation recurrence:
$c^{(n+1)}_i = \frac{i}{n+1} c^{(n)}_{i-1} + \left(1 - \frac{i}{n+1}\right) c^{(n)}_i$.
The surrogate \gls{gp} is fit from scratch on the lifted coefficients.
This method only supports $d=1$.

\FloatBarrier
\section{Convergence of the Integral Functionals}
\label{app:convergence}

\begin{figure}[h]
    \centering
    \includegraphics[width=\textwidth]{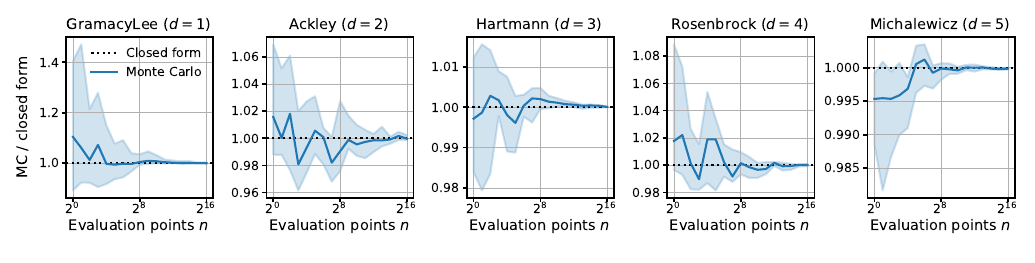}
    \caption{Ratio between the Monte Carlo estimate of the ridge functionals on $n$ uniform points and their closed form value (median and 95\% \gls{ci} over $32$ random candidates).}
    \label{fig:convergence}
\end{figure}

The ridge functionals of \cref{sec:experiments} are defined by the integrals $L_i[\vu] = \int w_i \vu$.
For candidates in the \gls{rbf} \gls{rkhs}, each integral factorizes over the input dimensions and each factor is a Gaussian integral of a trigonometric function over $[0,1]$, which we evaluate in closed form through the complex error function.
To check this closed form, and to show how accurately the functional can be approximated when it is not available, we compare it with the Monte Carlo estimate obtained by averaging $w_i \vu$ over $n$ uniform points in $[0,1]^d$.
We draw $32$ random candidates with $10$ support points and lengthscale $0.2$, in the same way as the initial designs of the sweep, and evaluate the ratio between the Monte Carlo and closed form target values for $n = 2^0, \ldots, 2^{16}$.
\Cref{fig:convergence} shows that the ratio converges to one on every benchmark: the median relative error is below $1.3\%$ at $n=2^{12}$ and below $0.3\%$ at $n=2^{16}$, decreasing at the expected $O(1/\sqrt{n})$ rate.

\FloatBarrier
\section{Wall-Clock Time}
\label{app:timing}

We report two timing studies.
\Cref{tab:timings_single} reports the total time per run spent fitting the surrogate and optimizing the acquisitions, measured on a single CPU core so that all methods run on the same hardware, and averaged across benchmarks.
\wycoff{} sits in the middle of the pack: it is cheaper than \vien{}, whose candidates carry the support points of all previous observations, and than the subspace methods \kundu{} and \shilton{} in this setting, while \vellanky{} has the cheapest acquisition since it optimizes $k+1$ polynomial coefficients only.
The surrogate fit costs about the same for all \gls{rkhs}-based methods, since they share the same \gls{gp} on \gls{rkhs} distances.

\Cref{tab:timings_cluster} breaks the times down per benchmark, as recorded during the main sweep.
Since the sweep ran on a heterogeneous cluster, these times depend on the node each run was scheduled on and are comparable only within a row.
The acquisition cost of \wycoff{} grows with the number of optimized parameters $k(d+m)$, and is largest on Hopper, where the policy has two outputs, and on the expensive MNIST and Pinwheel benchmarks.

\begin{table}[h]
\centering
\caption{Total wall-clock time per run (seconds) spent fitting the surrogate and optimizing the acquisition, over the $110$-evaluation budget, on a single CPU core and averaged across benchmarks.}
\label{tab:timings_single}
\setlength{\tabcolsep}{2pt}
\renewcommand{\arraystretch}{1.0}
\tiny
\begin{tabular*}{\textwidth}{@{\extracolsep{\fill}} l rrrrrr}
\toprule
\textbf{Time (s)} & Random & \kundu & \vien & \shilton & \vellanky & \wycoff \\
\midrule
Surrogate fit & $\text{\ding{55}}$ & $85.5$ & $88.5$ & $81.1$ & $154.1$ & $86.9$ \\
Acquisition & $3.9$ & $254.9$ & $586.1$ & $891.7$ & $79.7$ & $202.2$ \\
\bottomrule
\end{tabular*}
\end{table}

\input{tables/timings_cluster}

\FloatBarrier
\newpage
\input{checklist.tex}

\end{document}

%% file: tables/results_combined.tex
\begin{table}[t]
\centering
\caption{Average and final regret (median and 95\% CI) across methods and benchmark functions. Columns group projections (left), functional ridges (middle) and real-world tasks (right). $\text{\ding{55}}$ indicates the method does not support the corresponding problem class.}
\label{tab:regret_combined}
\setlength{\tabcolsep}{2pt}
\renewcommand{\arraystretch}{1.0}
\tiny
\begin{tabular}{l cc cc cc}
\toprule
& \multicolumn{2}{c}{\textbf{Sinc-1D} [$10^{-2}$]} & \multicolumn{2}{c}{\textbf{GramacyLee} [$10^{-1}$]} & \multicolumn{2}{c}{\textbf{Brachistochrone} [$10^{-2}$]} \\
\cmidrule(lr){2-3} \cmidrule(lr){4-5} \cmidrule(lr){6-7}
\textbf{Method} & \textbf{Avg} & \textbf{Final} & \textbf{Avg} & \textbf{Final} & \textbf{Avg} & \textbf{Final} \\
\midrule
Random     & $8.02\ [6.26, 8.89]$ & $6.67\ [5.01, 7.43]$ & $5.53\ [5.38, 5.61]$ & $5.19\ [5.14, 5.24]$ & $12.4\ [11.8, 13.0]$ & $9.80\ [8.06, 11.5]$ \\
\kundu     & $8.48\ [7.79, 8.69]$ & $2.38\ [2.06, 2.76]$ & $6.56\ [6.50, 6.65]$ & $6.37\ [6.37, 6.37]$ & $10.8\ [10.3, 11.3]$ & $5.16\ [4.86, 5.51]$ \\
\vien      & $2.58\ [2.31, 2.82]$ & $0.31\ [0.26, 0.33]$ & $5.45\ [5.39, 5.53]$ & $5.13\ [5.13, 5.13]$ & $7.42\ [7.15, 7.83]$ & $3.28\ [3.11, 3.52]$ \\
\shilton   & $6.86\ [6.16, 7.72]$ & $0.19\ [0.18, 0.26]$ & $5.50\ [5.38, 5.63]$ & $5.13\ [5.13, 5.13]$ & $5.07\ [3.91, 6.78]$ & $\mathbf{0.87\ [0.84, 0.91]}$ \\
\vellanky  & $9.65\ [9.62, 9.73]$ & $6.24\ [6.10, 6.46]$ & $5.52\ [5.41, 6.48]$ & $5.13\ [5.13, 6.37]$ & $9.15\ [8.98, 9.42]$ & $2.26\ [2.25, 2.29]$ \\
\wycoff    & $\mathbf{1.93\ [1.85, 2.04]}$ & $\mathbf{0.06\ [0.05, 0.07]}$ & $\mathbf{4.35\ [4.01, 5.04]}$ & $\mathbf{3.42\ [2.08, 4.35]}$ & $\mathbf{4.71\ [4.46, 4.83]}$ & $1.02\ [1.01, 1.06]$ \\
\midrule\midrule
& \multicolumn{2}{c}{\textbf{Sinc-2D} [$10^{-2}$]} & \multicolumn{2}{c}{\textbf{Ackley} [$10^{1}$]} & \multicolumn{2}{c}{\textbf{MNIST} [$10^{-2}$]} \\
\cmidrule(lr){2-3} \cmidrule(lr){4-5} \cmidrule(lr){6-7}
\textbf{Method} & \textbf{Avg} & \textbf{Final} & \textbf{Avg} & \textbf{Final} & \textbf{Avg} & \textbf{Final} \\
\midrule
Random     & $6.80\ [6.52, 7.20]$ & $6.38\ [6.01, 6.99]$ & $1.58\ [1.55, 1.59]$ & $1.46\ [1.43, 1.51]$ & $2.18\ [2.15, 2.19]$ & $2.16\ [2.14, 2.18]$ \\
\kundu     & $6.45\ [6.37, 6.55]$ & $3.85\ [3.64, 4.01]$ & $1.60\ [1.58, 1.61]$ & $1.51\ [1.49, 1.53]$ & $2.25\ [2.23, 2.27]$ & $2.09\ [2.08, 2.10]$ \\
\vien      & $4.74\ [4.37, 5.06]$ & $2.02\ [1.82, 2.21]$ & $1.60\ [1.57, 1.62]$ & $1.54\ [1.45, 1.58]$ & $2.17\ [2.16, 2.20]$ & $2.15\ [2.12, 2.17]$ \\
\shilton   & $6.66\ [6.42, 7.04]$ & $3.56\ [3.45, 3.81]$ & $1.39\ [1.35, 1.41]$ & $1.15\ [1.07, 1.22]$ & $2.15\ [2.14, 2.20]$ & $2.09\ [2.07, 2.12]$ \\
\vellanky  & \text{\ding{55}} & \text{\ding{55}} & \text{\ding{55}} & \text{\ding{55}} & $2.36\ [2.23, 2.93]$ & $2.17\ [2.14, 2.21]$ \\
\wycoff    & $\mathbf{2.70\ [2.65, 2.81]}$ & $\mathbf{0.76\ [0.70, 0.80]}$ & $\mathbf{0.66\ [0.61, 0.84]}$ & $\mathbf{0.09\ [0.06, 0.25]}$ & $\mathbf{2.11\ [2.10, 2.12]}$ & $\mathbf{2.06\ [2.04, 2.08]}$ \\
\midrule\midrule
& \multicolumn{2}{c}{\textbf{Sinc-3D} [$10^{-2}$]} & \multicolumn{2}{c}{\textbf{Hartmann} [$10^{0}$]} & \multicolumn{2}{c}{\textbf{Pendulum} [$10^{2}$]} \\
\cmidrule(lr){2-3} \cmidrule(lr){4-5} \cmidrule(lr){6-7}
\textbf{Method} & \textbf{Avg} & \textbf{Final} & \textbf{Avg} & \textbf{Final} & \textbf{Avg} & \textbf{Final} \\
\midrule
Random     & $5.22\ [5.04, 5.41]$ & $4.99\ [4.76, 5.23]$ & $3.14\ [3.11, 3.16]$ & $3.04\ [3.00, 3.07]$ & $5.29\ [5.22, 5.31]$ & $4.99\ [4.94, 5.05]$ \\
\kundu     & $5.09\ [5.00, 5.19]$ & $3.59\ [3.53, 3.63]$ & $2.94\ [2.91, 2.95]$ & $2.65\ [2.63, 2.68]$ & $5.08\ [5.05, 5.13]$ & $4.91\ [4.88, 4.93]$ \\
\vien      & $3.96\ [3.78, 4.05]$ & $3.16\ [3.10, 3.25]$ & $2.56\ [2.54, 2.58]$ & $2.37\ [2.36, 2.40]$ & $5.05\ [5.01, 5.12]$ & $4.63\ [4.58, 4.72]$ \\
\shilton   & $5.46\ [5.12, 5.68]$ & $3.43\ [3.40, 3.48]$ & $2.58\ [2.52, 2.59]$ & $2.42\ [2.38, 2.46]$ & $4.74\ [4.72, 4.77]$ & $4.34\ [4.30, 4.40]$ \\
\vellanky  & \text{\ding{55}} & \text{\ding{55}} & \text{\ding{55}} & \text{\ding{55}} & \text{\ding{55}} & \text{\ding{55}} \\
\wycoff    & $\mathbf{3.40\ [3.36, 3.44]}$ & $\mathbf{2.74\ [2.69, 2.77]}$ & $\mathbf{1.36\ [1.35, 1.37]}$ & $\mathbf{0.78\ [0.78, 0.78]}$ & $\mathbf{4.22\ [4.12, 4.33]}$ & $\mathbf{3.36\ [3.27, 3.55]}$ \\
\midrule\midrule
& \multicolumn{2}{c}{\textbf{Sinc-4D} [$10^{-2}$]} & \multicolumn{2}{c}{\textbf{Rosenbrock} [$10^{4}$]} & \multicolumn{2}{c}{\textbf{Pinwheel} [$10^{0}$]} \\
\cmidrule(lr){2-3} \cmidrule(lr){4-5} \cmidrule(lr){6-7}
\textbf{Method} & \textbf{Avg} & \textbf{Final} & \textbf{Avg} & \textbf{Final} & \textbf{Avg} & \textbf{Final} \\
\midrule
Random     & $3.62\ [3.58, 3.75]$ & $3.46\ [3.33, 3.60]$ & $6.57\ [6.51, 6.63]$ & $6.10\ [5.92, 6.21]$ & $1.99\ [1.87, 2.03]$ & $1.23\ [1.08, 1.42]$ \\
\kundu     & $4.39\ [4.19, 4.54]$ & $2.64\ [2.62, 2.67]$ & $5.91\ [5.85, 5.94]$ & $5.33\ [5.31, 5.37]$ & $1.75\ [1.68, 1.83]$ & $1.47\ [1.42, 1.54]$ \\
\vien      & $2.73\ [2.68, 2.76]$ & $\mathbf{2.43\ [2.42, 2.44]}$ & $5.39\ [5.33, 5.49]$ & $4.75\ [4.68, 4.89]$ & $2.95\ [2.86, 3.08]$ & $2.25\ [2.19, 2.33]$ \\
\shilton   & $5.37\ [4.79, 6.00]$ & $2.60\ [2.58, 2.63]$ & $5.05\ [4.97, 5.13]$ & $4.42\ [4.30, 4.46]$ & $1.73\ [1.69, 1.77]$ & $1.50\ [1.48, 1.59]$ \\
\vellanky  & \text{\ding{55}} & \text{\ding{55}} & \text{\ding{55}} & \text{\ding{55}} & $2.80\ [2.79, 2.81]$ & $2.53\ [2.53, 2.54]$ \\
\wycoff    & $\mathbf{2.73\ [2.70, 2.79]}$ & $2.47\ [2.46, 2.48]$ & $\mathbf{3.04\ [2.90, 3.20]}$ & $\mathbf{0.20\ [0.19, 0.22]}$ & $\mathbf{1.07\ [1.02, 1.22]}$ & $\mathbf{0.12\ [0.10, 0.18]}$ \\
\midrule\midrule
& \multicolumn{2}{c}{\textbf{Sinc-5D} [$10^{-2}$]} & \multicolumn{2}{c}{\textbf{Michalewicz} [$10^{0}$]} & \multicolumn{2}{c}{\textbf{Hopper} [$-10^{0}$]} \\
\cmidrule(lr){2-3} \cmidrule(lr){4-5} \cmidrule(lr){6-7}
\textbf{Method} & \textbf{Avg} & \textbf{Final} & \textbf{Avg} & \textbf{Final} & \textbf{Avg} & \textbf{Final} \\
\midrule
Random     & $4.59\ [4.49, 4.72]$ & $4.45\ [4.26, 4.59]$ & $4.20\ [4.18, 4.21]$ & $4.11\ [4.09, 4.14]$ & $1.84\ [1.75, 1.95]$ & $2.36\ [2.14, 2.51]$ \\
\kundu     & $4.85\ [4.75, 4.92]$ & $3.62\ [3.60, 3.66]$ & $3.96\ [3.93, 3.97]$ & $3.72\ [3.70, 3.75]$ & $1.87\ [1.79, 1.98]$ & $1.92\ [1.82, 2.02]$ \\
\vien      & $3.68\ [3.64, 3.71]$ & $\mathbf{3.36\ [3.36, 3.37]}$ & $3.74\ [3.72, 3.75]$ & $3.38\ [3.34, 3.42]$ & $1.53\ [1.50, 1.54]$ & $1.59\ [1.55, 1.63]$ \\
\shilton   & $6.55\ [6.05, 6.96]$ & $3.42\ [3.40, 3.43]$ & $3.81\ [3.79, 3.82]$ & $3.65\ [3.63, 3.67]$ & $2.23\ [2.19, 2.28]$ & $2.53\ [2.43, 2.57]$ \\
\vellanky  & \text{\ding{55}} & \text{\ding{55}} & \text{\ding{55}} & \text{\ding{55}} & $1.46\ [1.46, 1.46]$ & $1.47\ [1.47, 1.47]$ \\
\wycoff    & $\mathbf{3.67\ [3.63, 3.70]}$ & $3.40\ [3.38, 3.43]$ & $\mathbf{3.22\ [3.21, 3.22]}$ & $\mathbf{2.93\ [2.89, 2.93]}$ & $\mathbf{2.36\ [2.28, 2.53]}$ & $\mathbf{2.96\ [2.69, 3.05]}$ \\
\bottomrule
\end{tabular}
\end{table}

%% file: tables/best_lengthscales.tex
\begin{table}[h]
\centering
\caption{Selected hyperparameters per method and benchmark. The \gls{gp} kernel determines the order of differentiability $\mathcal{C}(\nu)$ of the surrogate model ($\mathcal{C}(\infty)$ = RBF). The \gls{rkhs} lengthscale $\ell$ is shared by all methods on a benchmark.}
\label{tab:hyperparams}
\setlength{\tabcolsep}{2pt}
\renewcommand{\arraystretch}{1.0}
\tiny
\begin{tabular*}{\textwidth}{@{\extracolsep{\fill}} l c cccc}
\toprule
\textbf{Benchmark} & \textbf{\gls{rkhs} scale} & \multicolumn{4}{c}{\textbf{Surrogate kernel}} \\
\cmidrule(lr){3-6}
 & & \kundu & \vien & \shilton & \wycoff \\
\midrule
GramacyLee & $0.4$ & $\mathcal{C}(\infty)$ & $\mathcal{C}(1)$ & $\mathcal{C}(2)$ & $\mathcal{C}(0)$ \\
Ackley & $0.4$ & $\mathcal{C}(2)$ & $\mathcal{C}(0)$ & $\mathcal{C}(1)$ & $\mathcal{C}(2)$ \\
Hartmann & $0.4$ & $\mathcal{C}(1)$ & $\mathcal{C}(\infty)$ & $\mathcal{C}(\infty)$ & $\mathcal{C}(2)$ \\
Rosenbrock & $0.4$ & $\mathcal{C}(2)$ & $\mathcal{C}(\infty)$ & $\mathcal{C}(\infty)$ & $\mathcal{C}(1)$ \\
Michalewicz & $0.4$ & $\mathcal{C}(\infty)$ & $\mathcal{C}(2)$ & $\mathcal{C}(2)$ & $\mathcal{C}(\infty)$ \\
\midrule
Brachistochrone & $0.2$ & $\mathcal{C}(0)$ & $\mathcal{C}(1)$ & $\mathcal{C}(1)$ & $\mathcal{C}(2)$ \\
MNIST & $0.4$ & $\mathcal{C}(0)$ & $\mathcal{C}(1)$ & $\mathcal{C}(0)$ & $\mathcal{C}(0)$ \\
Pendulum & $0.2$ & $\mathcal{C}(1)$ & $\mathcal{C}(0)$ & $\mathcal{C}(2)$ & $\mathcal{C}(1)$ \\
Pinwheel & $0.2$ & $\mathcal{C}(1)$ & $\mathcal{C}(0)$ & $\mathcal{C}(1)$ & $\mathcal{C}(2)$ \\
Hopper & $0.1$ & $\mathcal{C}(0)$ & $\mathcal{C}(\infty)$ & $\mathcal{C}(\infty)$ & $\mathcal{C}(1)$ \\
\midrule
Sinc-1D & $0.05$ & $\mathcal{C}(\infty)$ & $\mathcal{C}(2)$ & $\mathcal{C}(\infty)$ & $\mathcal{C}(1)$ \\
Sinc-2D & $0.1$ & $\mathcal{C}(\infty)$ & $\mathcal{C}(\infty)$ & $\mathcal{C}(\infty)$ & $\mathcal{C}(\infty)$ \\
Sinc-3D & $0.2$ & $\mathcal{C}(1)$ & $\mathcal{C}(2)$ & $\mathcal{C}(1)$ & $\mathcal{C}(\infty)$ \\
Sinc-4D & $0.4$ & $\mathcal{C}(1)$ & $\mathcal{C}(2)$ & $\mathcal{C}(0)$ & $\mathcal{C}(\infty)$ \\
Sinc-5D & $0.4$ & $\mathcal{C}(1)$ & $\mathcal{C}(\infty)$ & $\mathcal{C}(0)$ & $\mathcal{C}(\infty)$ \\
\bottomrule
\end{tabular*}
\end{table}

%% file: tables/lengthscale_ranks.tex
\begin{table}[h]
\centering
\caption{Mean rank over the ten main benchmarks at each \gls{rkhs} lengthscale $\ell$, each method using its own best \gls{gp} profile at that $\ell$. A method ties with any better one whose median lies within its own 95\% CI. Lower is better; best per row in bold.}
\label{tab:lengthscale_ranks}
\setlength{\tabcolsep}{2pt}
\renewcommand{\arraystretch}{1.0}
\tiny
\begin{tabular*}{\textwidth}{@{\extracolsep{\fill}} l cc cc cc cc cc cc }
\toprule
$\ell$ & \multicolumn{2}{c}{Random} & \multicolumn{2}{c}{\kundu} & \multicolumn{2}{c}{\vien} & \multicolumn{2}{c}{\shilton} & \multicolumn{2}{c}{\vellanky} & \multicolumn{2}{c}{\wycoff} \\
\cmidrule(lr){2-3} \cmidrule(lr){4-5} \cmidrule(lr){6-7} \cmidrule(lr){8-9} \cmidrule(lr){10-11} \cmidrule(lr){12-13}
 & \textbf{Avg} & \textbf{Final} & \textbf{Avg} & \textbf{Final} & \textbf{Avg} & \textbf{Final} & \textbf{Avg} & \textbf{Final} & \textbf{Avg} & \textbf{Final} & \textbf{Avg} & \textbf{Final} \\
\midrule
$0.05$ & $4.40$ & $4.70$ & $3.70$ & $3.20$ & $2.80$ & $2.30$ & $2.20$ & $2.90$ & $3.20$ & $3.20$ & $\mathbf{2.10}$ & $\mathbf{1.70}$ \\
$0.1$ & $4.20$ & $4.40$ & $3.40$ & $3.30$ & $3.10$ & $2.90$ & $2.60$ & $2.60$ & $3.20$ & $3.20$ & $\mathbf{1.70}$ & $\mathbf{1.60}$ \\
$0.2$ & $4.30$ & $4.40$ & $3.80$ & $3.60$ & $3.10$ & $3.40$ & $2.20$ & $2.50$ & $3.40$ & $3.40$ & $\mathbf{1.10}$ & $\mathbf{1.10}$ \\
$0.4$ & $3.50$ & $3.80$ & $4.10$ & $4.20$ & $3.10$ & $3.50$ & $1.90$ & $2.40$ & $4.60$ & $4.00$ & $\mathbf{1.10}$ & $\mathbf{1.10}$ \\
\bottomrule
\end{tabular*}
\end{table}

%% file: tables/lengthscale_full.tex
\begin{table}[h]
\centering
\caption{Median final and average regret at every \gls{rkhs} lengthscale, each method at its own best \gls{gp} profile. Each benchmark carries a common power of ten. Methods tied for first (see \cref{tab:lengthscale_ranks}) in bold.}
\label{tab:lengthscale_full}
\setlength{\tabcolsep}{2pt}
\renewcommand{\arraystretch}{1.0}
\tiny
\begin{tabular*}{\textwidth}{@{\extracolsep{\fill}} l cc cc cc cc cc cc }
\toprule
\textbf{Benchmark} & \multicolumn{2}{c}{Random} & \multicolumn{2}{c}{\kundu} & \multicolumn{2}{c}{\vien} & \multicolumn{2}{c}{\shilton} & \multicolumn{2}{c}{\vellanky} & \multicolumn{2}{c}{\wycoff} \\
\cmidrule(lr){2-3} \cmidrule(lr){4-5} \cmidrule(lr){6-7} \cmidrule(lr){8-9} \cmidrule(lr){10-11} \cmidrule(lr){12-13}
 & \textbf{Avg} & \textbf{Final} & \textbf{Avg} & \textbf{Final} & \textbf{Avg} & \textbf{Final} & \textbf{Avg} & \textbf{Final} & \textbf{Avg} & \textbf{Final} & \textbf{Avg} & \textbf{Final} \\
\midrule
\multicolumn{13}{l}{\textit{\gls{rkhs} lengthscale $\ell = 0.05$}} \\
GramacyLee [$10^{-1}$] & $7.30$ & $6.37$ & $6.53$ & $6.37$ & $6.90$ & $6.37$ & $6.46$ & $6.37$ & $\mathbf{5.52}$ & $\mathbf{5.13}$ & $6.89$ & $6.37$ \\
Ackley [$10^{1}$] & $1.69$ & $1.68$ & $1.69$ & $1.68$ & $\mathbf{1.69}$ & $1.68$ & $\mathbf{1.69}$ & $\mathbf{1.66}$ & \text{\ding{55}} & \text{\ding{55}} & $\mathbf{1.69}$ & $1.68$ \\
Hartmann [$10^{0}$] & $3.48$ & $3.48$ & $3.48$ & $3.47$ & $3.47$ & $\mathbf{3.46}$ & $\mathbf{3.47}$ & $3.47$ & \text{\ding{55}} & \text{\ding{55}} & $\mathbf{3.47}$ & $\mathbf{3.46}$ \\
Rosenbrock [$10^{4}$] & $7.87$ & $7.86$ & $7.86$ & $7.86$ & $\mathbf{7.86}$ & $\mathbf{7.84}$ & $\mathbf{7.86}$ & $7.85$ & \text{\ding{55}} & \text{\ding{55}} & $7.86$ & $7.84$ \\
Michalewicz [$10^{0}$] & $4.37$ & $4.37$ & $4.37$ & $4.37$ & $4.37$ & $\mathbf{4.37}$ & $\mathbf{4.37}$ & $4.37$ & \text{\ding{55}} & \text{\ding{55}} & $4.37$ & $\mathbf{4.37}$ \\
Brachistochrone [$10^{-1}$] & $2.66$ & $2.51$ & $3.03$ & $2.16$ & $1.86$ & $1.15$ & $3.53$ & $2.83$ & $0.91$ & $0.23$ & $\mathbf{0.85}$ & $\mathbf{0.19}$ \\
MNIST [$10^{-2}$] & $2.47$ & $2.45$ & $2.82$ & $2.36$ & $2.36$ & $2.34$ & $4.01$ & $2.48$ & $\mathbf{2.34}$ & $\mathbf{2.17}$ & $2.40$ & $2.36$ \\
Pendulum [$10^{2}$] & $5.48$ & $5.42$ & $5.34$ & $5.21$ & $5.37$ & $5.23$ & $5.31$ & $5.20$ & \text{\ding{55}} & \text{\ding{55}} & $\mathbf{5.18}$ & $\mathbf{4.87}$ \\
Pinwheel [$10^{0}$] & $1.82$ & $1.17$ & $\mathbf{0.60}$ & $\mathbf{0.41}$ & $1.95$ & $1.64$ & $\mathbf{0.61}$ & $0.57$ & $2.80$ & $2.53$ & $1.78$ & $1.16$ \\
Hopper [$-10^{0}$] & $1.73$ & $2.09$ & $1.88$ & $2.05$ & $1.64$ & $2.07$ & $\mathbf{2.16}$ & $2.37$ & $1.46$ & $1.47$ & $\mathbf{2.12}$ & $\mathbf{2.43}$ \\
\textit{mean rank} & $4.40$ & $4.70$ & $3.70$ & $3.20$ & $2.80$ & $2.30$ & $2.20$ & $2.90$ & $3.20$ & $3.20$ & $\mathbf{2.10}$ & $\mathbf{1.70}$ \\
\midrule\midrule
\multicolumn{13}{l}{\textit{\gls{rkhs} lengthscale $\ell = 0.1$}} \\
GramacyLee [$10^{-1}$] & $6.38$ & $5.41$ & $6.57$ & $6.37$ & $5.89$ & $5.13$ & $5.79$ & $5.13$ & $\mathbf{5.49}$ & $\mathbf{5.13}$ & $6.48$ & $6.37$ \\
Ackley [$10^{1}$] & $1.68$ & $1.65$ & $1.62$ & $\mathbf{1.50}$ & $1.69$ & $1.67$ & $\mathbf{1.58}$ & $\mathbf{1.49}$ & \text{\ding{55}} & \text{\ding{55}} & $\mathbf{1.63}$ & $\mathbf{1.49}$ \\
Hartmann [$10^{0}$] & $3.46$ & $3.45$ & $3.43$ & $3.39$ & $3.43$ & $3.37$ & $3.42$ & $3.39$ & \text{\ding{55}} & \text{\ding{55}} & $\mathbf{3.34}$ & $\mathbf{3.20}$ \\
Rosenbrock [$10^{4}$] & $7.81$ & $7.77$ & $7.73$ & $7.65$ & $7.69$ & $7.54$ & $7.69$ & $7.60$ & \text{\ding{55}} & \text{\ding{55}} & $\mathbf{7.54}$ & $\mathbf{7.25}$ \\
Michalewicz [$10^{0}$] & $4.37$ & $4.37$ & $4.36$ & $4.36$ & $4.36$ & $4.36$ & $4.36$ & $4.36$ & \text{\ding{55}} & \text{\ding{55}} & $\mathbf{4.36}$ & $\mathbf{4.35}$ \\
Brachistochrone [$10^{-1}$] & $2.04$ & $1.83$ & $1.79$ & $0.74$ & $1.24$ & $0.64$ & $2.28$ & $1.39$ & $0.91$ & $0.23$ & $\mathbf{0.69}$ & $\mathbf{0.10}$ \\
MNIST [$10^{-2}$] & $2.43$ & $2.42$ & $2.68$ & $2.31$ & $\mathbf{2.31}$ & $2.29$ & $2.95$ & $2.31$ & $\mathbf{2.42}$ & $\mathbf{2.17}$ & $2.38$ & $2.31$ \\
Pendulum [$10^{2}$] & $5.36$ & $5.21$ & $5.09$ & $4.80$ & $5.16$ & $4.78$ & $4.99$ & $4.77$ & \text{\ding{55}} & \text{\ding{55}} & $\mathbf{4.49}$ & $\mathbf{3.46}$ \\
Pinwheel [$10^{0}$] & $1.75$ & $\mathbf{0.99}$ & $\mathbf{0.92}$ & $\mathbf{0.88}$ & $2.25$ & $1.90$ & $1.10$ & $1.01$ & $2.80$ & $2.53$ & $1.75$ & $0.94$ \\
Hopper [$-10^{0}$] & $1.84$ & $2.36$ & $1.87$ & $1.92$ & $1.53$ & $1.59$ & $2.23$ & $2.53$ & $1.46$ & $1.47$ & $\mathbf{2.36}$ & $\mathbf{2.96}$ \\
\textit{mean rank} & $4.20$ & $4.40$ & $3.40$ & $3.30$ & $3.10$ & $2.90$ & $2.60$ & $2.60$ & $3.20$ & $3.20$ & $\mathbf{1.70}$ & $\mathbf{1.60}$ \\
\midrule\midrule
\multicolumn{13}{l}{\textit{\gls{rkhs} lengthscale $\ell = 0.2$}} \\
GramacyLee [$10^{-1}$] & $6.08$ & $5.33$ & $6.48$ & $6.37$ & $6.34$ & $5.18$ & $\mathbf{5.48}$ & $\mathbf{5.13}$ & $\mathbf{5.52}$ & $\mathbf{5.13}$ & $\mathbf{5.79}$ & $\mathbf{5.13}$ \\
Ackley [$10^{1}$] & $1.64$ & $1.57$ & $1.56$ & $1.43$ & $1.60$ & $1.49$ & $\mathbf{1.40}$ & $\mathbf{1.18}$ & \text{\ding{55}} & \text{\ding{55}} & $1.60$ & $\mathbf{1.52}$ \\
Hartmann [$10^{0}$] & $3.39$ & $3.33$ & $3.19$ & $2.96$ & $3.03$ & $2.65$ & $3.09$ & $2.98$ & \text{\ding{55}} & \text{\ding{55}} & $\mathbf{2.27}$ & $\mathbf{1.06}$ \\
Rosenbrock [$10^{4}$] & $7.48$ & $7.24$ & $6.76$ & $6.04$ & $6.64$ & $5.57$ & $6.67$ & $6.24$ & \text{\ding{55}} & \text{\ding{55}} & $\mathbf{4.58}$ & $\mathbf{2.01}$ \\
Michalewicz [$10^{0}$] & $4.34$ & $4.32$ & $4.28$ & $4.21$ & $4.28$ & $4.16$ & $4.27$ & $4.23$ & \text{\ding{55}} & \text{\ding{55}} & $\mathbf{3.99}$ & $\mathbf{3.67}$ \\
Brachistochrone [$10^{-1}$] & $1.24$ & $0.98$ & $1.08$ & $0.52$ & $0.74$ & $0.33$ & $\mathbf{0.51}$ & $\mathbf{0.09}$ & $0.91$ & $0.23$ & $\mathbf{0.47}$ & $0.10$ \\
MNIST [$10^{-2}$] & $2.31$ & $2.30$ & $2.46$ & $2.21$ & $2.27$ & $2.24$ & $2.37$ & $\mathbf{2.16}$ & $2.42$ & $\mathbf{2.17}$ & $\mathbf{2.22}$ & $\mathbf{2.16}$ \\
Pendulum [$10^{2}$] & $5.29$ & $4.99$ & $5.08$ & $4.91$ & $5.05$ & $4.63$ & $4.74$ & $4.34$ & \text{\ding{55}} & \text{\ding{55}} & $\mathbf{4.22}$ & $\mathbf{3.36}$ \\
Pinwheel [$10^{0}$] & $1.99$ & $1.23$ & $1.75$ & $1.47$ & $2.95$ & $2.25$ & $1.73$ & $1.50$ & $2.80$ & $2.53$ & $\mathbf{1.07}$ & $\mathbf{0.12}$ \\
Hopper [$-10^{0}$] & $\mathbf{1.64}$ & $\mathbf{1.87}$ & $1.54$ & $1.55$ & $1.46$ & $1.50$ & $1.62$ & $1.72$ & $1.46$ & $1.47$ & $\mathbf{1.69}$ & $\mathbf{1.77}$ \\
\textit{mean rank} & $4.30$ & $4.40$ & $3.80$ & $3.60$ & $3.10$ & $3.40$ & $2.20$ & $2.50$ & $3.40$ & $3.40$ & $\mathbf{1.10}$ & $\mathbf{1.10}$ \\
\midrule\midrule
\multicolumn{13}{l}{\textit{\gls{rkhs} lengthscale $\ell = 0.4$}} \\
GramacyLee [$10^{-1}$] & $5.53$ & $5.19$ & $6.56$ & $6.37$ & $5.45$ & $5.13$ & $5.50$ & $5.13$ & $5.52$ & $5.13$ & $\mathbf{4.35}$ & $\mathbf{3.42}$ \\
Ackley [$10^{1}$] & $1.58$ & $1.46$ & $1.60$ & $1.51$ & $1.60$ & $1.54$ & $1.39$ & $1.15$ & \text{\ding{55}} & \text{\ding{55}} & $\mathbf{0.66}$ & $\mathbf{0.09}$ \\
Hartmann [$10^{0}$] & $3.14$ & $3.04$ & $2.94$ & $2.65$ & $2.56$ & $2.37$ & $2.58$ & $2.42$ & \text{\ding{55}} & \text{\ding{55}} & $\mathbf{1.36}$ & $\mathbf{0.78}$ \\
Rosenbrock [$10^{4}$] & $6.57$ & $6.10$ & $5.91$ & $5.33$ & $5.39$ & $4.75$ & $5.05$ & $4.42$ & \text{\ding{55}} & \text{\ding{55}} & $\mathbf{3.04}$ & $\mathbf{0.20}$ \\
Michalewicz [$10^{0}$] & $4.20$ & $4.11$ & $3.96$ & $3.72$ & $3.74$ & $3.38$ & $3.81$ & $3.65$ & \text{\ding{55}} & \text{\ding{55}} & $\mathbf{3.22}$ & $\mathbf{2.93}$ \\
Brachistochrone [$10^{-2}$] & $7.65$ & $3.64$ & $8.68$ & $7.06$ & $7.16$ & $5.41$ & $\mathbf{3.10}$ & $\mathbf{1.05}$ & $9.15$ & $2.26$ & $4.22$ & $1.25$ \\
MNIST [$10^{-2}$] & $2.18$ & $2.16$ & $2.25$ & $2.09$ & $2.17$ & $2.15$ & $2.15$ & $2.09$ & $2.36$ & $2.17$ & $\mathbf{2.11}$ & $\mathbf{2.06}$ \\
Pendulum [$10^{2}$] & $5.27$ & $5.01$ & $5.27$ & $5.17$ & $5.17$ & $5.05$ & $\mathbf{4.71}$ & $\mathbf{4.28}$ & \text{\ding{55}} & \text{\ding{55}} & $\mathbf{4.77}$ & $\mathbf{4.31}$ \\
Pinwheel [$10^{0}$] & $2.34$ & $1.88$ & $2.33$ & $1.99$ & $3.04$ & $2.69$ & $2.26$ & $2.05$ & $2.80$ & $2.53$ & $\mathbf{1.58}$ & $\mathbf{0.60}$ \\
Hopper [$-10^{0}$] & $1.56$ & $1.66$ & $1.46$ & $1.48$ & $1.47$ & $1.50$ & $1.53$ & $1.56$ & $1.46$ & $1.47$ & $\mathbf{1.67}$ & $\mathbf{1.76}$ \\
\textit{mean rank} & $3.50$ & $3.80$ & $4.10$ & $4.20$ & $3.10$ & $3.50$ & $1.90$ & $2.40$ & $4.60$ & $4.00$ & $\mathbf{1.10}$ & $\mathbf{1.10}$ \\
\bottomrule
\end{tabular*}
\end{table}

%% file: tables/matern52.tex
\begin{table}[h]
\centering
\caption{Median final and average regret with the Mat\'ern-5/2 \gls{gp} profile for every method, at the selected \gls{rkhs} lengthscale. Methods tied for first in bold.}
\label{tab:matern52}
\setlength{\tabcolsep}{2pt}
\renewcommand{\arraystretch}{1.0}
\tiny
\begin{tabular*}{\textwidth}{@{\extracolsep{\fill}} l cc cc cc cc cc cc }
\toprule
\textbf{Benchmark} & \multicolumn{2}{c}{Random} & \multicolumn{2}{c}{\kundu} & \multicolumn{2}{c}{\vien} & \multicolumn{2}{c}{\shilton} & \multicolumn{2}{c}{\vellanky} & \multicolumn{2}{c}{\wycoff} \\
\cmidrule(lr){2-3} \cmidrule(lr){4-5} \cmidrule(lr){6-7} \cmidrule(lr){8-9} \cmidrule(lr){10-11} \cmidrule(lr){12-13}
 & \textbf{Avg} & \textbf{Final} & \textbf{Avg} & \textbf{Final} & \textbf{Avg} & \textbf{Final} & \textbf{Avg} & \textbf{Final} & \textbf{Avg} & \textbf{Final} & \textbf{Avg} & \textbf{Final} \\
\midrule
GramacyLee [$10^{-1}$] & $\mathbf{5.53}$ & $5.19$ & $6.58$ & $6.37$ & $\mathbf{5.45}$ & $5.13$ & $\mathbf{5.50}$ & $\mathbf{5.13}$ & $5.46$ & $\mathbf{5.13}$ & $\mathbf{5.39}$ & $\mathbf{5.13}$ \\
Ackley [$10^{1}$] & $1.58$ & $1.46$ & $1.60$ & $1.51$ & $1.61$ & $1.58$ & $1.39$ & $1.16$ & \text{\ding{55}} & \text{\ding{55}} & $\mathbf{0.66}$ & $\mathbf{0.09}$ \\
Hartmann [$10^{0}$] & $3.14$ & $3.04$ & $2.94$ & $2.65$ & $2.55$ & $2.38$ & $2.58$ & $2.43$ & \text{\ding{55}} & \text{\ding{55}} & $\mathbf{1.36}$ & $\mathbf{0.78}$ \\
Rosenbrock [$10^{4}$] & $6.57$ & $6.10$ & $5.91$ & $5.33$ & $5.51$ & $5.10$ & $5.07$ & $4.43$ & \text{\ding{55}} & \text{\ding{55}} & $\mathbf{3.17}$ & $\mathbf{0.66}$ \\
Michalewicz [$10^{0}$] & $4.20$ & $4.11$ & $3.96$ & $3.72$ & $3.74$ & $3.38$ & $3.81$ & $3.65$ & \text{\ding{55}} & \text{\ding{55}} & $\mathbf{3.24}$ & $\mathbf{2.93}$ \\
Brachistochrone [$10^{-1}$] & $1.24$ & $0.98$ & $1.21$ & $0.74$ & $0.74$ & $0.38$ & $\mathbf{0.52}$ & $\mathbf{0.09}$ & $0.91$ & $0.23$ & $\mathbf{0.47}$ & $0.10$ \\
MNIST [$10^{-2}$] & $2.18$ & $2.16$ & $2.28$ & $2.19$ & $2.18$ & $2.15$ & $2.18$ & $\mathbf{2.14}$ & $2.34$ & $2.18$ & $\mathbf{2.15}$ & $\mathbf{2.13}$ \\
Pendulum [$10^{2}$] & $5.29$ & $4.99$ & $5.09$ & $4.95$ & $5.15$ & $4.75$ & $4.74$ & $4.34$ & \text{\ding{55}} & \text{\ding{55}} & $\mathbf{4.31}$ & $\mathbf{3.43}$ \\
Pinwheel [$10^{0}$] & $1.95$ & $1.26$ & $1.79$ & $1.55$ & $3.17$ & $2.78$ & $1.72$ & $1.54$ & $2.82$ & $2.54$ & $\mathbf{1.07}$ & $\mathbf{0.12}$ \\
Hopper [$-10^{0}$] & $1.84$ & $2.36$ & $1.86$ & $1.89$ & $1.50$ & $1.54$ & $\mathbf{2.23}$ & $\mathbf{2.51}$ & $1.46$ & $1.47$ & $\mathbf{2.16}$ & $\mathbf{2.53}$ \\
\midrule
\textit{mean rank} & $3.80$ & $4.00$ & $4.00$ & $4.30$ & $3.10$ & $3.50$ & $1.80$ & $1.80$ & $4.40$ & $3.80$ & $\mathbf{1.00}$ & $\mathbf{1.10}$ \\
\bottomrule
\end{tabular*}
\end{table}

%% file: tables/timings_cluster.tex
\begin{table}[h]
\centering
\caption{Median total wall-clock time per run (seconds) spent fitting the surrogate and optimizing the acquisition, over the $110$-evaluation budget, as recorded in the main sweep on a heterogeneous cluster. Times include \textsc{jax} compilation and are only indicative across rows.}
\label{tab:timings_cluster}
\setlength{\tabcolsep}{2pt}
\renewcommand{\arraystretch}{1.0}
\tiny
\begin{tabular*}{\textwidth}{@{\extracolsep{\fill}} l rr rr rr rr rr }
\toprule
\textbf{Benchmark} & \multicolumn{2}{c}{\kundu} & \multicolumn{2}{c}{\vien} & \multicolumn{2}{c}{\shilton} & \multicolumn{2}{c}{\vellanky} & \multicolumn{2}{c}{\wycoff} \\
\cmidrule(lr){2-3} \cmidrule(lr){4-5} \cmidrule(lr){6-7} \cmidrule(lr){8-9} \cmidrule(lr){10-11}
 & \textbf{Fit} & \textbf{Acq.} & \textbf{Fit} & \textbf{Acq.} & \textbf{Fit} & \textbf{Acq.} & \textbf{Fit} & \textbf{Acq.} & \textbf{Fit} & \textbf{Acq.} \\
\midrule
GramacyLee & $66$ & $191$ & $58$ & $496$ & $49$ & $71$ & $108$ & $72$ & $86$ & $78$ \\
Ackley & $87$ & $171$ & $63$ & $455$ & $45$ & $30$ & \text{\ding{55}} & \text{\ding{55}} & $99$ & $102$ \\
Hartmann & $69$ & $120$ & $69$ & $437$ & $67$ & $42$ & \text{\ding{55}} & \text{\ding{55}} & $82$ & $89$ \\
Rosenbrock & $96$ & $174$ & $74$ & $503$ & $72$ & $36$ & \text{\ding{55}} & \text{\ding{55}} & $98$ & $129$ \\
Michalewicz & $67$ & $173$ & $88$ & $756$ & $80$ & $46$ & \text{\ding{55}} & \text{\ding{55}} & $92$ & $234$ \\
Brachistochrone & $75$ & $291$ & $86$ & $563$ & $438$ & $1262$ & $160$ & $70$ & $90$ & $134$ \\
MNIST & $15362$ & $1392$ & $66$ & $386$ & $442$ & $1266$ & $5469$ & $175$ & $1320$ & $1671$ \\
Pendulum & $83$ & $182$ & $99$ & $560$ & $79$ & $31$ & \text{\ding{55}} & \text{\ding{55}} & $94$ & $143$ \\
Pinwheel & $2040$ & $1155$ & $81$ & $509$ & $424$ & $1116$ & $820$ & $205$ & $1246$ & $1487$ \\
Hopper & $81$ & $621$ & $102$ & $1238$ & $105$ & $671$ & $165$ & $79$ & $94$ & $1247$ \\
Sinc-1D & $88$ & $213$ & $76$ & $559$ & $104$ & $59$ & $177$ & $67$ & $84$ & $126$ \\
Sinc-2D & $91$ & $160$ & $103$ & $522$ & $96$ & $52$ & \text{\ding{55}} & \text{\ding{55}} & $104$ & $121$ \\
Sinc-3D & $106$ & $189$ & $97$ & $460$ & $106$ & $56$ & \text{\ding{55}} & \text{\ding{55}} & $63$ & $109$ \\
Sinc-4D & $82$ & $183$ & $101$ & $623$ & $89$ & $62$ & \text{\ding{55}} & \text{\ding{55}} & $67$ & $137$ \\
Sinc-5D & $75$ & $152$ & $102$ & $570$ & $86$ & $162$ & \text{\ding{55}} & \text{\ding{55}} & $91$ & $156$ \\
\bottomrule
\end{tabular*}
\end{table}

%% file: checklist.tex
\section*{NeurIPS Paper Checklist}

\begin{enumerate}

\item {\bf Claims}
    \item[] Question: Do the main claims made in the abstract and introduction accurately reflect the paper's contributions and scope?
    \item[] Answer: \answerYes{} 
    \item[] Justification: We mention our methodology and experimental claims in abstract and substantiate.

\item {\bf Limitations}
    \item[] Question: Does the paper discuss the limitations of the work performed by the authors?
    \item[] Answer: \answerYes{} 
    \item[] Justification: The Limitations paragraph of \cref{sec:discussion} covers the lack of convergence or regret guarantees, the non-convex acquisition problem, the restriction to objectives that are well approximated by few support points, the shared complexity schedule, the coarse hyperparameter grid and the computational cost, with pointers to the corresponding appendices.

\item {\bf Theory assumptions and proofs}
    \item[] Question: For each theoretical result, does the paper provide the full set of assumptions and a complete (and correct) proof?
    \item[] Answer: \answerYes{} 
    \item[] Justification: The paper is primarily methodological. Its formal statements (the structure of the sparse set as a finite union of immersed manifolds, density of its union in $\Hs$, existence of a maximizer of the acquisition problem, and the equivalence of the updates of prior work with natural gradient descent) are stated with their assumptions and derived in \cref{sec:methodology}. We do not claim convergence or regret guarantees.

    \item {\bf Experimental result reproducibility}
    \item[] Question: Does the paper fully disclose all the information needed to reproduce the main experimental results of the paper to the extent that it affects the main claims and/or conclusions of the paper (regardless of whether the code and data are provided or not)?
    \item[] Answer: \answerYes{} 
    \item[] Justification: We give details of numerical experiments in Section 5 and in appendices.

\item {\bf Open access to data and code}
    \item[] Question: Does the paper provide open access to the data and code, with sufficient instructions to faithfully reproduce the main experimental results, as described in supplemental material?
    \item[] Answer: \answerYes{} 
    \item[] Justification: The code, including the benchmark implementations and the script reproducing the full experimental sweep, is available at \url{https://github.com/davidesartor/bofus}. All data used are either generated by the benchmarks or publicly available (MNIST).

\item {\bf Experimental setting/details}
    \item[] Question: Does the paper specify all the training and test details (e.g., data splits, hyperparameters, how they were chosen, type of optimizer) necessary to understand the results?
    \item[] Answer: \answerYes{} 
    \item[] Justification: Yes we discuss this in the experimental section and in the appendices.

\item {\bf Experiment statistical significance}
    \item[] Question: Does the paper report error bars suitably and correctly defined or other appropriate information about the statistical significance of the experiments?
    \item[] Answer: \answerYes{} 
    \item[] Justification: All figures and tables report the median over 16 to 32 independent seeds together with a 95\% confidence interval for the median, computed from the order statistics at quantiles $0.5 \pm 1.96\sqrt{0.25/n}$ (normal approximation to the binomial), which does not assume normally distributed outcomes.

\item {\bf Experiments compute resources}
    \item[] Question: For each experiment, does the paper provide sufficient information on the computer resources (type of compute workers, memory, time of execution) needed to reproduce the experiments?
    \item[] Answer: \answerYes{} 
    \item[] Justification: \Cref{app:implementation} describes the sweep (about 105k CPU runs on a Slurm cluster) and \cref{app:timing} reports wall-clock times per run and per method, both on a single CPU core and per benchmark as recorded during the sweep. The repository provides the scripts to reproduce it with preset time and memory requirements.
    
\item {\bf Code of ethics}
    \item[] Question: Does the research conducted in the paper conform, in every respect, with the NeurIPS Code of Ethics \url{https://neurips.cc/public/EthicsGuidelines}?
    \item[] Answer: \answerYes{} 
    \item[] Justification: We followed that code.

\item {\bf Broader impacts}
    \item[] Question: Does the paper discuss both potential positive societal impacts and negative societal impacts of the work performed?
    \item[] Answer: \answerNA{} 
    \item[] Justification: This is foundational research on optimization algorithms, not tied to particular applications, and we do not foresee a direct societal impact beyond that of black-box optimization in general.
    
\item {\bf Safeguards}
    \item[] Question: Does the paper describe safeguards that have been put in place for responsible release of data or models that have a high risk for misuse (e.g., pre-trained language models, image generators, or scraped datasets)?
    \item[] Answer: \answerNA{} 
    \item[] Justification: Not applicable to this kind of research.

\item {\bf Licenses for existing assets}
    \item[] Question: Are the creators or original owners of assets (e.g., code, data, models), used in the paper, properly credited and are the license and terms of use explicitly mentioned and properly respected?
    \item[] Answer: \answerYes{} 
    \item[] Justification: We re-implemented the methods of prior work from scratch and cite them. The existing assets we use are cited: MNIST~\citep{lecun1998mnist} (CC BY-SA 3.0), the Pendulum environment of Gym/Gymnasium~\citep{brockman2016openai} (MIT license), \textsc{jax}~\citep{jax2018github} (Apache 2.0), and the benchmark functions of the virtual library of simulation experiments~\citep{surjanovic2013virtual}, which we re-implemented from their published definitions.

\item {\bf New assets}
    \item[] Question: Are new assets introduced in the paper well documented and is the documentation provided alongside the assets?
    \item[] Answer: \answerYes{} 
    \item[] Justification: Our code is released at \url{https://github.com/davidesartor/bofus} under the MIT license, with a README documenting installation, the methods and the benchmarks.

\item {\bf Crowdsourcing and research with human subjects}
    \item[] Question: For crowdsourcing experiments and research with human subjects, does the paper include the full text of instructions given to participants and screenshots, if applicable, as well as details about compensation (if any)? 
    \item[] Answer: \answerNA{} 
    \item[] Justification: No crowdsourcing or human subjects research performed.

\item {\bf Institutional review board (IRB) approvals or equivalent for research with human subjects}
    \item[] Question: Does the paper describe potential risks incurred by study participants, whether such risks were disclosed to the subjects, and whether Institutional Review Board (IRB) approvals (or an equivalent approval/review based on the requirements of your country or institution) were obtained?
    \item[] Answer: \answerNA{} 
    \item[] Justification: No human research conducted.

\item {\bf Declaration of LLM usage}
    \item[] Question: Does the paper describe the usage of LLMs if it is an important, original, or non-standard component of the core methods in this research? Note that if the LLM is used only for writing, editing, or formatting purposes and does \emph{not} impact the core methodology, scientific rigor, or originality of the research, declaration is not required.
    \item[] Answer: \answerNo{} 
    \item[] Justification: LLMs were not used for the core methods of research.

\end{enumerate}

%% file: references.bib
@article{hermus2021exploiting,
  title={Exploiting redundancy to facilitate physical interaction},
  author={Hermus, James and Lachner, Johannes and Verdi, David and Hogan, Neville},
  journal={IEEE Transactions on Robotics},
  volume={38},
  number={1},
  pages={599--615},
  year={2021},
  publisher={IEEE}
}

@inproceedings{vien2018bayesian,
  title={Bayesian functional optimization},
  author={Vien, Ngo Anh and Zimmermann, Heiko and Toussaint, Marc},
  booktitle={Proceedings of the AAAI Conference on Artificial Intelligence},
  volume={32},
  number={1},
  year={2018}
}

@article{posa2014direct,
  title={A direct method for trajectory optimization of rigid bodies through contact},
  author={Posa, Michael and Cantu, Cecilia and Tedrake, Russ},
  journal={The International Journal of Robotics Research},
  volume={33},
  number={1},
  pages={69--81},
  year={2014},
  publisher={Sage Publications Sage UK: London, England}
}

@inproceedings{vellanki2019bayesian,
  title={Bayesian functional optimisation with shape prior},
  author={Vellanki, Pratibha and Rana, Santu and Gupta, Sunil and de Celis Leal, David Rubin and Sutti, Alessandra and Height, Murray and Venkatesh, Svetha},
  booktitle={Proceedings of the AAAI Conference on Artificial Intelligence},
  volume={33},
  number={01},
  pages={1617--1624},
  year={2019}
}

@article{kundu2025active,
  title={Active Subspaces in Infinite Dimension},
  author={Kundu, Poorbita and Wycoff, Nathan},
  journal={arXiv preprint arXiv:2510.11871},
  year={2025}
}

@article{shilton2020sequential,
  title={Sequential subspace search for functional bayesian optimization incorporating experimenter intuition},
  author={Shilton, Alistair and Gupta, Sunil and Rana, Santu and Venkatesh, Svetha},
  journal={arXiv preprint arXiv:2009.03543},
  year={2020}
}

@article{amari1998natural,
  title={Natural gradient works efficiently in learning},
  author={Amari, Shun-Ichi},
  journal={Neural computation},
  volume={10},
  number={2},
  pages={251--276},
  year={1998},
  publisher={MIT Press}
}

@article{vincent2002kernel,
  title={Kernel matching pursuit},
  author={Vincent, Pascal and Bengio, Yoshua},
  journal={Machine learning},
  volume={48},
  number={1},
  pages={165--187},
  year={2002},
  publisher={Springer}
}

@article{shields2021bayesian,
  year = {2021},
  title = {Bayesian reaction optimization as a tool for chemical synthesis},
  author = {Shields, Benjamin J and Stevens, Jason and Li, Jun and Parasram, Marvin and Damani, Farhan and Alvarado, Jesus I Martinez and Janey, Jacob M and Adams, Ryan P and Doyle, Abigail G},
  journal = {Nature},
  volume = {590},
  number = {7844},
  pages = {89--96},
  publisher = {Nature Publishing Group UK London}
}

@article{frazier2015materials,
  title = {Bayesian optimization for materials design},
  author = {Frazier, Peter I. and Wang, Jialei},
  journal = {arXiv preprint arXiv:1506.01349},
  year = {2015}
}

@article{pedersen2021bayesian,
  title = {Bayesian Optimization of High-Entropy Alloy Compositions for Electrocatalytic Oxygen Reduction},
  author = {Pedersen, Jack K. and Clausen, Christian M. and Krysiak, Olga A. and Xiao, Bin and Batchelor, Thomas A. A. and Loffler, Tobias and Mints, Vladislav A. and Banko, Lars and Arenz, Matthias and Savan, Alan and Schuhmann, Wolfgang and Ludwig, Alfred and Rossmeisl, Jan},
  journal = {arXiv preprint arXiv:2106.08212},
  year = {2021}
}

@article{todorovic2019bayesian,
  title={Bayesian inference of atomistic structure in functional materials},
  author={Todorovi{\'c}, Milica and Gutmann, Michael U and Corander, Jukka and Rinke, Patrick},
  journal={Npj computational materials},
  volume={5},
  number={1},
  pages={35},
  year={2019},
  publisher={Nature Publishing Group UK London}
}

@article{wang2022bayesian,
  title={Bayesian optimization for chemical products and functional materials},
  author={Wang, Ke and Dowling, Alexander W},
  journal={Current Opinion in Chemical Engineering},
  volume={36},
  pages={100728},
  year={2022},
  publisher={Elsevier}
}

@article{wang2016rembo,
  title     = {Bayesian Optimization in a Billion Dimensions via Random Embeddings},
  author    = {Wang, Ziyu and Hutter, Frank and Zoghi, Masrour and Matheson, David and de Freitas, Nando},
  journal   = {Journal of Artificial Intelligence Research},
  volume    = {55},
  pages     = {361--387},
  year      = {2016},
  url       = {https://arxiv.org/abs/1301.1942}
}

@inproceedings{kandasamy2015additive,
  title     = {High Dimensional {B}ayesian Optimisation and Bandits via Additive Models},
  author    = {Kandasamy, Kirthevasan and Schneider, Jeff and P{\'o}czos, Barnab{\'a}s},
  booktitle = {Proceedings of the 32nd International Conference on Machine Learning},
  pages     = {295--304},
  year      = {2015},
  editor    = {Bach, Francis and Blei, David},
  volume    = {37},
  series    = {Proceedings of Machine Learning Research},
  publisher = {PMLR},
  url       = {https://proceedings.mlr.press/v37/kandasamy15.html}
}

@inproceedings{rolland2018additive,
  title     = {High-Dimensional {B}ayesian Optimization via Additive Models with Overlapping Groups},
  author    = {Rolland, Paul and Scarlett, Jonathan and Bogunovic, Ilija and Cevher, Volkan},
  booktitle = {Proceedings of the Twenty-First International Conference on Artificial Intelligence and Statistics (AISTATS)},
  pages     = {298--307},
  year      = {2018},
  volume    = {84},
  series    = {Proceedings of Machine Learning Research},
  publisher = {PMLR},
  url       = {https://proceedings.mlr.press/v84/rolland18a.html}
}

@inproceedings{han2021tree,
  title     = {High-Dimensional {B}ayesian Optimization via Tree-Structured Additive Models},
  author    = {Han, Eric and Arora, Ishank and Scarlett, Jonathan},
  booktitle = {Proceedings of the AAAI Conference on Artificial Intelligence},
  volume    = {35},
  number    = {9},
  pages     = {7630--7638},
  year      = {2021},
  doi       = {10.1609/aaai.v35i9.16933},
  url       = {https://ojs.aaai.org/index.php/AAAI/article/view/16933}
}

@inproceedings{papenmeier2022baxus,
  title     = {Increasing the Scope as You Learn: Adaptive {B}ayesian Optimization in Nested Subspaces},
  author    = {Papenmeier, Leonard and Nardi, Luigi and Poloczek, Matthias},
  booktitle = {Advances in Neural Information Processing Systems},
  volume    = {35},
  pages     = {11586--11601},
  year      = {2022},
  publisher = {Curran Associates, Inc.},
  url       = {https://openreview.net/pdf?id=e4Wf6112DI}
}

@inproceedings{eriksson2021saasbo,
  title     = {High-Dimensional {B}ayesian Optimization with Sparse Axis-Aligned Subspaces},
  author    = {Eriksson, David and Jankowiak, Martin},
  booktitle = {Proceedings of the Thirty-Seventh Conference on Uncertainty in Artificial Intelligence},
  pages     = {493--503},
  year      = {2021},
  editor    = {de Campos, Cassio and Maathuis, Marloes H.},
  volume    = {161},
  series    = {Proceedings of Machine Learning Research},
  publisher = {PMLR},
  url       = {https://proceedings.mlr.press/v161/eriksson21a.html}
}

@inproceedings{letham2020alebo,
  title     = {Re-Examining Linear Embeddings for High-Dimensional {B}ayesian Optimization},
  author    = {Letham, Benjamin and Calandra, Roberto and Rai, Akshara and Bakshy, Eytan},
  booktitle = {Advances in Neural Information Processing Systems},
  volume    = {33},
  pages     = {1546--1558},
  year      = {2020},
  url       = {https://arxiv.org/abs/2001.11659}
}

@inproceedings{nayebi2019hesbo,
  title     = {A Framework for {B}ayesian Optimization in Embedded Subspaces},
  author    = {Nayebi, Amin and Munteanu, Alexander and Poloczek, Matthias},
  booktitle = {Proceedings of the 36th International Conference on Machine Learning},
  pages     = {4752--4761},
  year      = {2019},
  editor    = {Chaudhuri, Kamalika and Salakhutdinov, Ruslan},
  volume    = {97},
  series    = {Proceedings of Machine Learning Research},
  publisher = {PMLR},
  url       = {https://proceedings.mlr.press/v97/nayebi19a.html}
}

@inproceedings{hvarfner2024vanilla,
  title     = {Vanilla {B}ayesian Optimization Performs Great in High Dimensions},
  author    = {Hvarfner, Carl and Hellsten, Erik Orm and Nardi, Luigi},
  booktitle = {Proceedings of the 41st International Conference on Machine Learning},
  pages     = {20793--20817},
  year      = {2024},
  editor    = {Salakhutdinov, Ruslan and Kolter, Zico and Heller, Katherine and Weller, Adrian and Oliver, Nuria and Scarlett, Jonathan and Berkenkamp, Felix},
  volume    = {235},
  series    = {Proceedings of Machine Learning Research},
  publisher = {PMLR},
  url       = {https://proceedings.mlr.press/v235/hvarfner24a.html}
}

@article{doumont2025we,
  title={We Still Don't Understand High-Dimensional Bayesian Optimization},
  author={Doumont, Colin and Fan, Donney and Maus, Natalie and Gardner, Jacob R and Moss, Henry and Pleiss, Geoff},
  journal={arXiv preprint arXiv:2512.00170},
  year={2025}
}

@book{rnw,
    author = {Rasmussen, Carl Edward and Williams, Christopher K. I.},
    title = {Gaussian Processes for Machine Learning},
    publisher = {The MIT Press},
    year = {2005},
    month = {11},
    isbn = {9780262256834},
    doi = {10.7551/mitpress/3206.001.0001},
    url = {https://doi.org/10.7551/mitpress/3206.001.0001},
    eprint = {https://direct.mit.edu/book-pdf/2514321/book_9780262256834.pdf},
}

@article{muehlenstaedt2017computer,
  title={Computer experiments with functional inputs and scalar outputs by a norm-based approach},
  author={Muehlenstaedt, Thomas and Fruth, Jana and Roustant, Olivier},
  journal={Statistics and Computing},
  volume={27},
  number={4},
  pages={1083--1097},
  year={2017},
  publisher={Springer}
}

@misc{betancourt2020fungp,
  title={fungp: Gaussian process models for scalar and functional inputs},
  author={Betancourt, J and Bachoc, F and Klein, T and Rohmer, J},
  year={2020},
  publisher={R package version 0.3. 0, URL https://cran. r-project. org/package= funGp}
}

@article{damiano2022automatic,
  title={Automatic Dynamic Relevance Determination for Gaussian process regression with high-dimensional functional inputs},
  author={Damiano, Luis and Johnson, Margaret and Teixeira, Joaquim and Morris, Max D and Niemi, Jarad},
  journal={arXiv preprint arXiv:2209.00044},
  year={2022}
}

@inproceedings{kirschner2019adaptive,
  title={Adaptive and safe Bayesian optimization in high dimensions via one-dimensional subspaces},
  author={Kirschner, Johannes and Mutny, Mojmir and Hiller, Nicole and Ischebeck, Rasmus and Krause, Andreas},
  booktitle={International Conference on Machine Learning},
  pages={3429--3438},
  year={2019},
  organization={PMLR}
}

@software{jax2018github,
  author = {James Bradbury and Roy Frostig and Peter Hawkins and Matthew James Johnson and Yash Katariya and Chris Leary and Dougal Maclaurin and George Necula and Adam Paszke and Jake Vander{P}las and Skye Wanderman-{M}ilne and Qiao Zhang},
  title = {{JAX}: composable transformations of {P}ython+{N}um{P}y programs},
  url = {http://github.com/jax-ml/jax},
  version = {0.3.13},
  year = {2018},
}

@article{jones1998efficient,
  title={Efficient global optimization of expensive black-box functions},
  author={Jones, Donald R and Schonlau, Matthias and Welch, William J},
  journal={Journal of Global Optimization},
  volume={13},
  number={4},
  pages={455--492},
  year={1998},
  publisher={Springer}
}

@article{frazier2018tutorial,
  title={A tutorial on {B}ayesian optimization},
  author={Frazier, Peter I.},
  journal={arXiv preprint arXiv:1807.02811},
  year={2018}
}

@book{garnett2023bayesian,
  author={Garnett, Roman},
  title={{B}ayesian Optimization},
  publisher={Cambridge University Press},
  year={2023}
}

@article{kanagawa2018gaussian,
  title={Gaussian processes and kernel methods: A review on connections and equivalences},
  author={Kanagawa, Motonobu and Hennig, Philipp and Sejdinovic, Dino and Sriperumbudur, Bharath K},
  journal={arXiv preprint arXiv:1807.02582},
  year={2018}
}

@misc{surjanovic2013virtual,
  author={Surjanovic, S. and Bingham, D.},
  title={Virtual Library of Simulation Experiments: Test Functions and Datasets},
  year={2013},
  note={\url{http://www.sfu.ca/~ssurjano}, retrieved \today}
}

@article{lecun1998mnist,
  title={Gradient-based learning applied to document recognition},
  author={LeCun, Yann and Bottou, L{\'e}on and Bengio, Yoshua and Haffner, Patrick},
  journal={Proceedings of the IEEE},
  volume={86},
  number={11},
  pages={2278--2324},
  year={1998}
}

@article{brockman2016openai,
  title={{O}pen{AI} {G}ym},
  author={Brockman, Greg and Cheung, Vicki and Pettersson, Ludwig and Schneider, Jonas and Schulman, John and Tang, Jie and Zaremba, Wojciech},
  journal={arXiv preprint arXiv:1606.01540},
  year={2016}
}

@article{ament2024logei,
  title={Unexpected improvements to expected improvement for {B}ayesian optimization},
  author={Ament, Sebastian and Daulton, Samuel and Eriksson, David and Balandat, Maximilian and Bakshy, Eytan},
  journal={Advances in Neural Information Processing Systems},
  volume={36},
  year={2024}
}

@article{byrd1995lbfgsb,
  title={A limited memory algorithm for bound constrained optimization},
  author={Byrd, Richard H and Lu, Peihuang and Nocedal, Jorge and Zhu, Ciyou},
  journal={SIAM Journal on Scientific Computing},
  volume={16},
  number={5},
  pages={1190--1208},
  year={1995},
  publisher={SIAM}
}

@article{kim2026validation,
  title={Validation of Dynamic Bayesian Optimization for a Non-Stationary Human-in-the-Loop Optimization Problem},
  author={Kim, GilHwan and Sergi, Fabrizio},
  journal={IEEE Robotics and Automation Letters},
  year={2026},
  publisher={IEEE}
}

@article{binois2021hetgp,
  title={hetgp: Heteroskedastic Gaussian process modeling and sequential design in R},
  author={Binois, Micka{\"e}l and Gramacy, Robert B},
  journal={Journal of Statistical Software},
  volume={98},
  pages={1--44},
  year={2021}
}

@incollection{forrester2008,
  author    = {Alexander I. J. Forrester and Andr{\'a}s S{\'o}bester and Andy J. Keane},
  title     = {Sampling Plans},
  booktitle = {Engineering Design via Surrogate Modelling: A Practical Guide},
  chapter   = {1},
  publisher = {John Wiley \& Sons, Ltd},
  year      = {2008},
}

@book{gramacy2020surrogates,
  title={Surrogates: Gaussian process modeling, design, and optimization for the applied sciences},
  author={Gramacy, Robert B},
  year={2020},
  publisher={Chapman and Hall/CRC}
}

@book{pinkus2015ridge,
  title={Ridge functions},
  author={Pinkus, Allan},
  volume={205},
  year={2015},
  publisher={Cambridge University Press}
}
